\documentclass{article}

\usepackage{colortbl}
\usepackage{xcolor}

\usepackage{microtype}
\usepackage{graphicx}
\usepackage{subcaption}
\usepackage{booktabs} 
\usepackage{colortbl}
\usepackage{hyperref}
\usepackage{url}
\usepackage{subcaption}
\usepackage{adjustbox}
\usepackage{needspace}
\usepackage[group-separator={,},group-minimum-digits = 4]{siunitx}
\usepackage{array}    
\usepackage{tcolorbox}
\usepackage{tabularx}
\usepackage{booktabs}
\usepackage{array}
\usepackage{makecell}
\usepackage{times}
\usepackage{enumitem}
\usepackage{latexsym}
\usepackage{float}
\usepackage{booktabs}
\usepackage{amsmath, amssymb}
\usepackage{hyperref}
\usepackage{xspace}
\usepackage{booktabs}
\usepackage{multirow}
\usepackage{longtable}
\usepackage[table]{xcolor}
\usepackage{array}
\usepackage{xcolor}
\usepackage{array}
\usepackage{wrapfig}
\usepackage{tabularx}
\usepackage{tcolorbox}
\usepackage[dvipsnames]{xcolor}
\definecolor{headercolor}{gray}{0.85}  %
\usepackage{inconsolata} % for better texttt

\usepackage{tikz}
\newcommand{\circled}[1]{\tikz[baseline=(char.base)]{
    \node[shape=circle,draw=black,fill=black,inner sep=1pt,minimum size=3mm] (char) {\textcolor{white}{\scriptsize\textbf{#1}}};}}

\usepackage{tikz}

\newcommand{\vanilla}{Original\xspace}
\newcommand{\segment}{Rephrasing\xspace}
\newcommand{\testcase}{Example Synthesis\xspace}
\newcommand{\shorthand}{Shorthand\xspace}
\newcommand{\samples}{examples\xspace}
\newcommand{\sample}{example\xspace}
\newcommand{\relevant}{\textit{Relevant}\xspace}
\newcommand{\partially}{\textit{Partially Relevant}\xspace}
\newcommand{\irrelevant}{\textit{Irrelevant}\xspace}
\newcommand{\recall}{$\text{Recall}@k$\xspace}
\newcommand{\ndcg}{$\text{NDCG}@k$\xspace}
\newcommand{\methodp}{$\text{Prec}_\text{
anchor}@k$\xspace}
\newcommand{\systemp}{$\text{Prec}_\text{
sys}@k$\xspace}
\newcommand{\usecase}{use-case\xspace}
\newcommand{\usecases}{use-cases\xspace}
\newcommand{\database}{benchmark database\xspace}

\newcommand{\topic}{\textit{topics} set\xspace}
\newcommand{\skill}{\textit{skills} set\xspace}
\newcommand{\application}{\textit{applications} set\xspace}
\newcommand{\novel}{\textit{novel} set\xspace}
\newcommand{\known}{\textit{known validation} set\xspace}

\newcommand{\methodname}{\textsc{BenchBrowser}\xspace}

\usepackage{array}    
\usepackage{tcolorbox}
\definecolor{lightyellow}{rgb}{1.0, 1.0, 0.88}

\usepackage{placeins}
\usepackage{array}
\usepackage{ragged2e}
\definecolor{lightpurple}{RGB}{230, 220, 255}
\definecolor{lightgreen}{RGB}{220, 255, 230}
\definecolor{darkpurple}{RGB}{120, 80, 160}
\definecolor{darkgreen}{RGB}{60, 140, 80}

\usepackage{tcolorbox}
\tcbuselibrary{skins, breakable}

\newtcolorbox{takeawaybox}[2][]{
    enhanced,
    breakable,
    colback=gray!15,         % background fill
    colframe=black,          % outer border
    colbacktitle=black,      % title background
    coltitle=white,          % title text color
    fonttitle=\bfseries,     % bold title font
    title={#2},              % title content
    left=8pt, right=8pt,
    top=6pt, bottom=6pt,
    boxrule=1pt,
    sharp corners,
    attach boxed title to top left={yshift=-1mm},
    boxed title style={
        sharp corners,
        boxrule=1pt,
    },
    #1
}

\definecolor{lightorange}{rgb}{1, 0.9, 0.8}
\definecolor{lightblue}{rgb}{0.8, 0.9, 1}

\newcolumntype{L}{>{\raggedright\arraybackslash}X}

\newcommand{\draftonly}[1]{#1}

\renewcommand{\draftonly}[1]{}

\definecolor{deeppurple}{HTML}{9e02f7}

\usepackage{floatflt}
\usepackage{tcolorbox}
\tcbset{
  sidebyside,
  sidebyside align=top,
  colback=white,
  colframe=white,
}

\usepackage{hyperref}

\usepackage[preprint]{icml2026}

\usepackage{amsmath}
\usepackage{amssymb}
\usepackage{mathtools}
\usepackage{amsthm}
\usepackage{bm}

\usepackage[capitalize,noabbrev]{cleveref}

\theoremstyle{plain}

\theoremstyle{definition}

\theoremstyle{remark}

\usepackage[textsize=tiny]{todonotes}

\icmltitlerunning{BenchBrowser: Retrieving Evidence for Evaluating Benchmark Validity}

\begin{document}

\twocolumn[
  \icmltitle{BenchBrowser: Retrieving Evidence for Evaluating Benchmark Validity}



  \icmlsetsymbol{equal}{*}

  \begin{icmlauthorlist}
    \icmlauthor{Harshita Diddee}{yyy}
    \icmlauthor{Gregory Yauney}{}
    \icmlauthor{Swabha Swayamdipta}{comp}
    \icmlauthor{Daphne Ippolito}{yyy}
  \end{icmlauthorlist}

  \icmlaffiliation{yyy}{Language Technologies Institute, Carnegie Mellon University, Pittsburgh (PA), United States}
  \icmlaffiliation{comp}{Thomas Lord Department of Computer Science, University of Southern California, Los Angeles (CA), United States}

  \icmlcorrespondingauthor{Harshita Diddee}{hdiddee@andrew.cmu.edu}

  \icmlkeywords{Machine Learning, ICML}

  \vskip 0.3in
]



\printAffiliationsAndNotice{}  % no special notice (required even if empty)

% \swabha{i could not find anything revealing here, but just wanted to double check that this is anonymized.}
% \diddee{I double checked too! Removing the paper link tab for posterity but no information seems identifying.}
\begin{abstract}
\textit{Do language model benchmarks actually measure what practitioners intend them to?}
High-level metadata is too coarse to convey the granular reality of benchmarks: a ``poetry" benchmark may never test for haikus, while ``instruction-following'' benchmarks will often test for an arbitrary mix of skills. This opacity makes verifying alignment with practitioner goals a laborious process, risking an illusion of competence even when models fail on untested facets of user interests. We introduce \methodname, a retriever that surfaces evaluation items relevant to natural-language use cases across 20+ benchmark suites. Validated by a human study confirming high retrieval precision, \methodname generates evidence to help practitioners diagnose low \textit{content validity} (narrow coverage of a capability's facets) and low \textit{convergent validity} (lack of stable rankings when measuring the same capability). \methodname helps quantify a critical gap between practitioner intent and what benchmarks actually test.
The \href{https://custombenchwebsite-802374347260.northamerica-northeast1.run.app}{tool} and its \href{https://github.com/harshitadd/BenchBrowser}{code} are publicly available.
\end{abstract}
\section{Introduction}
Practitioners often operate under a perilous assumption: that public benchmarks act as reliable proxies for their specific \usecases. Yet benchmarks differ widely in both scope and granularity.
For instance, in some benchmarks each example probes a different capability, as with InfoBench \citep{qin2024infobench} and AlpacaEval \citep{dubois2024length}. But in others, all examples test the same targeted capability, as in GSM8K \citep{cobbe2021training} and NutriBench \citep{Hua2024NutriBenchAD}. 
This heterogeneity in design can compromise the \emph{validity} of our evaluations, i.e., the degree to which the performance of a model on a benchmark reflects its true competence in the capability of interest \citep{Bean2025MeasuringWM, ye2025large}.
To confidently deploy a model, a practitioner must then answer a difficult question: \textit{Does this benchmark provide comprehensive evidence of a model's competence for my specific \usecase?}

\begin{figure}[t]
    \centering
    \includegraphics[width=0.95\linewidth]{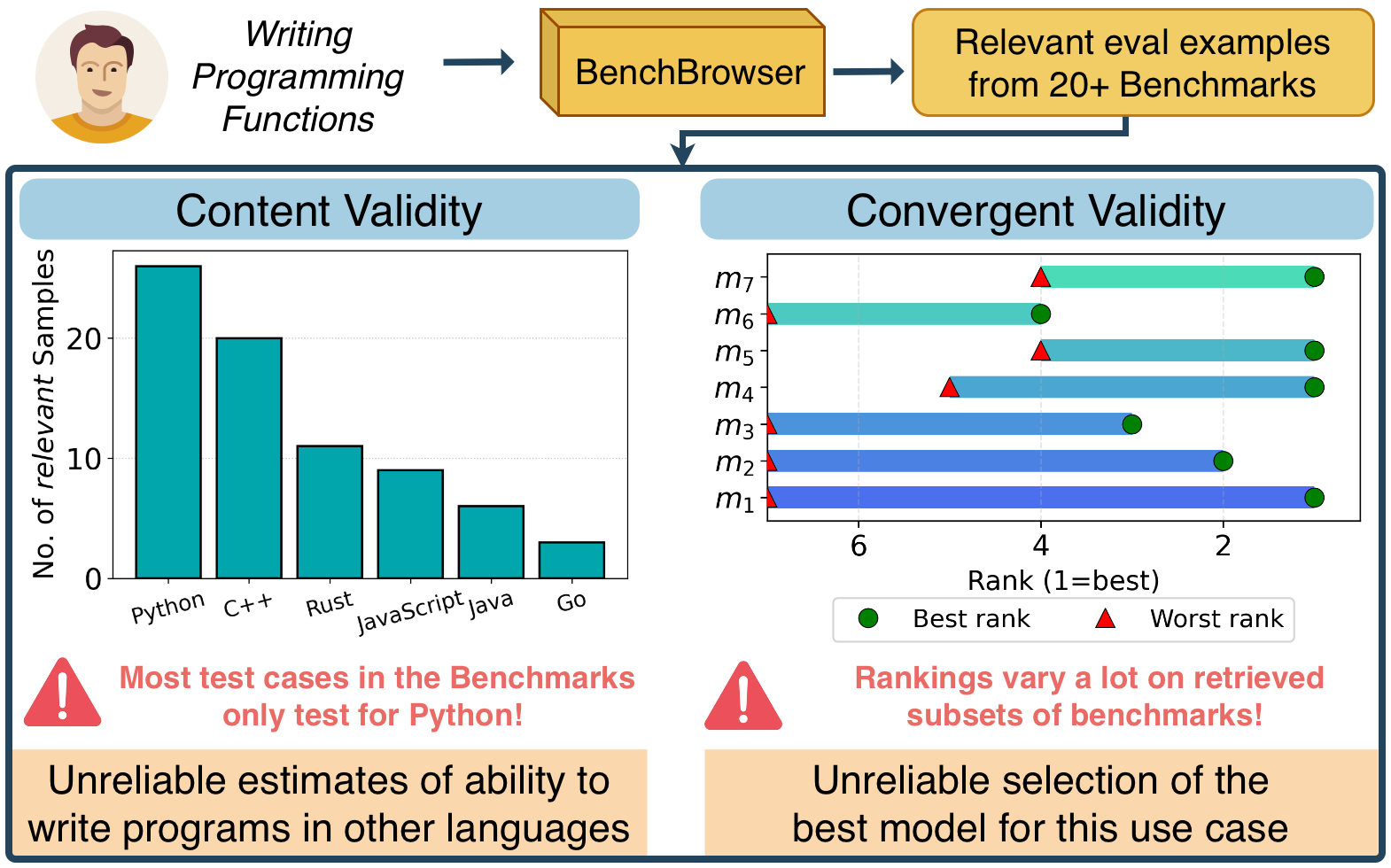}
    \caption{\methodname retrieves use-case-relevant items from 20+ benchmarks to help diagnose validity failures. For a ``Writing Programming Functions'' use-case: Python-skewed coverage (left; content validity gap) and unstable model ranks across subsets (right; low convergent validity) make conclusions unreliable. $m_{1}$ to $m_{k}$ are mid-sized decoder models. See Appendix~\ref{app:motivation-experiment} for details.}
    \label{fig:teaser}
    \vspace{-14pt}
\end{figure}

\begin{figure*}
    \centering
    \includegraphics[width=1\linewidth]{plots/zoomed.png}
    \caption{\methodname pipeline: a practitioner submits a \usecase of interest to \methodname; the \usecase is \circled{a} rewritten into anchors for more diverse retrieval, which are then \circled{b} embedded and used to retrieve \circled{c} relevant \samples from a suite of 20+ benchmarks.
    The retrieved evidence can be used by the practitioner in multiple ways (images show snapshots of \methodname's website UI): \circled{$d_{1}$} Comparing the representation of the \usecase along different facets. For example, a user interested in assessing ``reasoning skills'' can observe the different ways (or lack thereof) in which reasoning competence is evaluated (linguistics fundamentals versus math problems) and \circled{$d_{2}$} Compare the performance and stability of models evaluated on existing benchmarks (e.g., ARC AI2 Reasoning) for the task as well as subsets of the retrieved \samples.}
    \label{fig:system}
\end{figure*}

Answering this question is difficult because the gap between a benchmark's description and its content creates two distinct validity pitfalls. First, evaluations can lack \textit{content validity}: the extent to which the items in a test adequately represent the entire domain of the capability being measured. This can be compromised by \emph{sparse coverage}, where a practitioner probing for a multi-faceted capability finds few relevant evaluation items in a broad evaluation suite, or by \emph{scope-metadata mismatch}, where broad benchmark descriptions mislead. For example, BIG-bench's \texttt{Hindu\_Knowledge} task \citep{srivastava2023beyond} tests deity hierarchy but doesn't cover origins of spiritual texts---one of the many aspects that a practitioner might be interested in.
%which are topics a practitioner may assume are tested by a benchmark called ``Hindu Knowledge.'' 
Second, evaluations can often degrade \textit{convergent validity}: this occurs when models evaluated on different benchmarks measuring the same capability produce conflicting rankings. A model might excel at ``reasoning'' on science questions (AI2 ARC, \citealt{clark2018think}) but could fail to ``reason'' about consistencies in narrative summarization (FlawedFictions, \citealt{ahuja2025finding}). Confronted with these \emph{divergent operationalizations} of reasoning competence, the practitioner may be left unsure of which signal to trust to choose a competent model for their \usecase \citep{eriksson2025can}.

Diagnosing these failures is currently a prohibitively slow and expensive process. Existing methods such as manual auditing \citep{Bean2025MeasuringWM} or brittle keyword searches on dataset cards cannot scale to the tens of thousands of items in modern suites \citep{ma2025task}. Furthermore, validity is inherently stakeholder-dependent. One practitioner's definition of ``coding ability'' may focus on Python syntax, while another's might focus on Go concurrency. Static benchmark documentation cannot continuously capture these evolving definitions. To bridge this gap, we need tools that impose no assumptions about what a capability ``should'' mean, but instead empowers practitioners to retrieve operationalizations that match their own mental model \citep{wallach2025position}.

In this work, we introduce \methodname, a retrieval tool that helps practitioners quickly gather evidence of how benchmarks represent their \usecase (Figure~\ref{fig:teaser} and \S\ref{sec:method}).
Practitioners can query \methodname with a natural language  \usecase like ``\textit{ethical decision-making in AI art}'' or ``\textit{organic chemistry}'', and collect semantically related \samples across tens of thousands of \samples across popular benchmark suites in seconds. This speed enables interactive auditing, where practitioners can iteratively inspect which data points impact their \usecase's evaluation.

We validate the reliability of the evidence generated by \methodname in a user study with $N=10$ NLP practitioners, finding that \methodname returns \samples that practitioners deem relevant for diverse \usecases\ (\S\ref{sec:retrieval-eval}). 

We then use this evidence to diagnose two risks to validity.
First, we show how it can surface content validity failures, where minor facet changes expose stark representational disparities for \usecase variants (\S\ref{sec:content-validity}). For instance, we show how swapping a single facet in a use-case like ``programming in \underline{Python}'' to ``programming in \underline{Go}'' results in a sharp drop in the number of available \samples, demonstrating that competence estimates in ``programming'' may be disproportionately anchored to this specific language. 
Second, we show how evidence from \methodname can uncover convergent validity failures when retrieved \samples---even those judged as relevant by human annotators---induce model rankings that diverge from rankings derived from existing human-validated benchmarks for the same \usecase (\S\ref{sec:convergent-validity}). For example, when evaluating models on their ability to answer questions about the Hindu religion, the top-ranked models on the multiple-choice BIG-bench \texttt{Hindu\_Knowledge} task can drop to nearly the worst-ranked when instead evaluated on prompt-completion type \samples from HellaSwag \cite{zellers2019hellaswag}.

Overall, \methodname surfaces evidence for practitioners to directly assess validity rather than relying on rigid annotation schema, enabling interactive evaluation of benchmark validity for novel use-cases.

% \swabha{examples of concrete benchmarks and their convergent validity / content validity. }

% \swabha{make the human validation a focus.}

\section{\methodname}
\label{sec:method}

\methodname is a tool that retrieves \samples relevant to a natural-language \usecase (Figure~\ref{fig:system}). 
Given a \usecase and an integer $k$, we return the top-$k$ most relevant \samples from a large database spanning many benchmarks. 
Our method has three stages: (1) Anchor generation: rewrite the \usecase into one or more retrieval anchors; (2) Retrieval: embed anchors and fetch nearest \samples; (3) Filtering: an LLM judge removes obvious off-topic items.
\subsection{Stage 1: Anchor Generation}
\label{sec:retrieval-anchors}
%\swabha{how exactly do you achieve this? do u have the baseline format for every benchmark?}
% In theory, we do (I did check the format of one-sample per benchmark to influence the prompt for \testcase generation) but I did not include a comprehensive list of all the formats in the \testcase generation prompt.

Using practitioner \usecases directly for retrieval is insufficient: practitioner intents are often abstract (e.g., ``instruction following'') and benchmark \samples are rarely annotated for such subjective, fine-grained intents. Moreover, \usecases  omit format cues (e.g., multiple-choice vs.\ long-form) that are implicit in benchmark \samples, and such mismatches can reduce retrieval effectiveness \cite{noel1997influence}.
Query-rewriting is a standard technique to bridge this gap between queries and document representations  \cite{rizvi2004extending,10.1145/1367497.1367714,ma2023query}.
We test four methods for generating rewritten \textit{anchors} from queries, described below and summarized in Table~\ref{tab:anchor_examples}.
Full details for all methods are in Appendix~\ref{app:anchor-generation}.
% We use these rewritten queries, which we call \textit{anchors}, to retrieve diverse, relevant \samples from our \database.
 
\paragraph{\vanilla.}
As a no-rewrite baseline, we use the practitioner's \usecase as an anchor. 

\paragraph{\segment.}
Practitioner \usecases can range from being multifaceted, broad topics to abstract capabilities which can attract a noisier set of \samples with weak relevance \cite{krovetz1992lexical,chen2013question}. To make the anchors for such \usecases more specific, we first ask \texttt{gpt-5-mini-2025-08-07} to determine whether the practitioner’s \usecase is best interpreted as a topic or a capability. If it is a topic, the model decomposes it into $n=3$ related sub-topics; if it is a capability, the model lists $n=3$ skills needed to perform tasks associated with that capability.

\paragraph{\testcase.} 
To overcome the lack of format cues in a \usecase, we ask \texttt{gpt-5-mini-2025-08-07} to generate $n=3$ synthetic \samples that would be helpful in assessing the competence of AI models for the practitioner's \usecase. We explicitly guide the model to generate such \samples in diverse formats informed by our audit of the formats in which \samples are represented in our \database (long-form QA, MCQ, fill-in-the-blank). This rewrite aims to maximize the coverage of retrieved \samples by using an anchor that is in-distribution with the forms in the database.

\paragraph{\shorthand.}
Most benchmarks do not include \samples that explicitly include a natural language description of the capability or topic that the \sample is testing.
For example, WinoGrande \citep{sakaguchi2020winogrande} \samples do not explicitly state that they are testing ``pronoun resolution''. 
Since explicit annotations which include the practitioner's \usecase can provide a stronger chance of alignment, we rewrite the query into a bounded-vocabulary \emph{shorthand} representation which includes explicit annotation for these signals in the following form: $\langle\text{skill}\rangle \;\&\; \langle\text{key}_1\rangle \;\&\; \langle\text{key}_2\rangle$. 
Each \shorthand anchor includes the primary skill tested in the \sample in addition to most salient topical cues to avoid losing nuanced information from the \sample. Unlike other anchor methods that query raw \samples, we first convert all \samples in the \database to the same \shorthand form and run retrieval in this space. This yields (a) a rewrite of the practitioner \usecase into a vocabulary that is more in-distribution to the \database and (b) embeddings where \samples with the same skill or related skills cluster more tightly due to explicit skill annotations. We implement this cost-efficiently by finetuning a \texttt{Llama-3-8B-Instruct} model to rewrite the database at a nominal one-time cost.

% \swabha{is bm25 used for anchor generation? presumably u could use bm25 with any of these generated anchors? shouldn't this be under stage 2?} 
% Yes! Makes sense. Both are naive retrieval baselines. Not anchor generation methods. 

\begin{table}[t!]
\centering
\small
\caption{We compare four methods for producing retrieval anchors. Here we show how each handles an example practitioner \usecase: ``\textit{understanding geopolitical tensions in Southeast Asia}''.}
\label{tab:anchor_examples}
\vspace{2pt}
\begin{tabularx}{\columnwidth}{X}
\toprule
\bf \vanilla: \\
understanding geopolitical tensions in Southeast Asia \\
\midrule
\bf \segment: \\
South China Sea territorial disputes and maritime tensions • ASEAN diplomacy, regional security mechanisms, and conflict prevention • US–China strategic competition \\
\midrule
\bf \testcase: \\
Discuss the strategic factors driving contemporary geopolitical tensions in Southeast Asia • MCQ: Which factor most significantly contributes to territorial disputes in the South China Sea? [...] • Fill in the blank: ASEAN's principle of \_\_\_\_\_\_\_\_\_\_\_ often limits collective action [...] \\
\midrule
\bf \shorthand: \\
geopolitical\_analysis \& southeast\_asia \& international\_relations \\
\bottomrule
\end{tabularx}
\vspace{-15pt}
\end{table}

\subsection{Stage 2: Semantic Retrieval}
We generate semantic embeddings for query anchors using an in-context learning (ICL) embedding model, \texttt{baai/bge-en-icl} \cite{bge_embedding}.
We chose this model for its performance and speed after benchmarking a diverse set of embedding models, a neural re-ranker, an LLM embedder specifically designed for diverse RAG applications, and this ICL embedding model (details in Appendix~\ref{app:retrievers}).
Following this, we run an efficient similarity search using Faiss \citep{douze2024faiss} and retrieve the top-$k$ nearest neighbor \samples for each anchor. We then combine and sort these \samples by their similarity score and pick the top-$k$ for the final filtering stage. 
We also compare to BM25, a classical lexical retrieval method \citep{robertson2009probabilistic} and random retrieval to quantify gains using semantic retrieval. 

% BM25 is a classical lexical retrieval method which helps quantifies the gain from semantic retrieval over naive keyword matching which is both, fast and crucial for certain \usecases, such as the ones that include Named Entities.

% \paragraph{Random.} A naive baseline retrieves \samples without using the \usecase. 

%\swabha{i think this needs a little more elaboration, looks like appendix c6 has all the details (knowledge overlap etc.) that are important.}
% Added a substantiation of the dimensions across which relevance is judged, 

\subsection{Stage 3: Filtering}
We then use an LLM judge (\texttt{gpt-5-mini-2025-08-07}) to further filter for relevance. The prompt for the LLM judge is designed to assess if the \samples returned by \methodname are related to the intent of the user's \usecase (details in Appendix~\ref{app:selection_prompt}).
The judge assesses relevance along three dimensions:
(a) the extent of topical overlap between the \sample and the \usecase;
(b) the degree of skill overlap between the competencies required to solve the \sample and those required for the \usecase; and
(c) whether success on the \sample provides credible evidence of competence with respect to the \usecase.
This step aims to eliminate only the most obviously unrelated examples and not to outright select the relevant examples, as the latter is a non-trivial aspect to judge and is ultimately the practitioner's prerogative.

\section{Evaluation Setup}
\label{sec:evaluation-categories}

% \swabha{not sure how the rest of the section connects to this paragraph - are u outlining the properties of each eval set?}

% Yes, just an introduction line to describe what the reader can expect for this section. I took out the ''summarized below`` outside the bracket to highlight that this is the prefix to the sections outlined. 

We evaluate \methodname\ on five categories of \usecase queries, which we design to mirror the diverse intents of real-world practitioners (summarized in Table~\ref{tab:retriever-usecases}).
Full details and prompts are in Appendix~\ref{app:evaluation-setup}. The categories are:

\vspace{-4pt}
\paragraph{Topics.}
Captures broad domains (e.g., ``sports'', ``law'') central to traditional suites like MMLU \citep{hendrycks2020measuring} to reflect the intent of practitioners interested in auditing coverage of specific knowledge areas. We use 58 subjects from BBH \citep{suzgun2023challenging} and MMLU.

\paragraph{Skills.}
Targets capabilities spanning multiple domains, such as ``reasoning'' or ``creative writing.'' These queries represent intents for practitioners who design skill-focused evaluations or who  assess how well a single skill (e.g., “critical thinking”) is represented across multiple benchmarks. To construct this set, we start from the Clio collection of real-world AI use-cases \citep{tamkin2024clio} and use \texttt{claude-sonnet-4-5-20250929} to synthesize natural language use-case descriptions phrased as broad skills.
\paragraph{Applications.}
Focuses on the highly contextual, concrete tasks typically sought from deployed LLMs (e.g., ``studying for the SAT''). Unlike the abstract \skill, this set mirrors the specific user requests common in instruction-following benchmarks \citep{qin2024infobench, zhou2023instruction, wen2024benchmarking}. These are also derived from Clio via \texttt{claude-sonnet-4-5-20250929}.

\definecolor{nude}{HTML}{228B22}
\definecolor{lightgray}{gray}{0.9}
\newcommand{\grayhline}{\arrayrulecolor{lightgray}\hline\arrayrulecolor{black}}

\begin{table}[t!]
\centering
\footnotesize
\setlength{\tabcolsep}{2pt}
\caption{Use-case query sets for evaluating the retrieval quality of \methodname. Automatically synthesized sets derive from benchmarks; manually curated sets are hand-crafted for validation.}
\label{tab:retriever-usecases}
\vspace{-3pt}
\resizebox{\linewidth}{!}{%
\begin{tabular}{
>{\raggedright\arraybackslash}p{0.22\columnwidth}
>{\raggedright\arraybackslash}p{0.22\columnwidth}
>{\centering\arraybackslash}p{0.08\columnwidth}
>{\raggedright\arraybackslash}p{0.45\columnwidth}
}
\toprule
\textbf{Name} & \textbf{Origin} & $\bm{N}$ & \textbf{Examples} \\ 
\midrule
\multicolumn{4}{c}{\small\textit{Automatically synthesized}} \\
\midrule 
\textit{Topics} & BBH, MMLU & 58 & Marketing, Law, HS Maths \\ 
\textit{Skills} & Clio & 50 & creative writing, navigation \\ 
\textit{Applications} & Clio & 50 & Tips for Winter Gardening \\ 
\midrule
\multicolumn{4}{c}{\small\textit{Manually curated}} \\
\midrule
\textit{Known Valid.} & LM Eval & 20 & Riddles, Calorie Counting \\ 
\textit{Novel} & Crowdsourced & 21 & Matplotlib for Research \\ 
\bottomrule
\end{tabular}
}
\end{table}

\paragraph{Novel.}
Because practitioners may also wish to audit new \usecases that are sparsely represented or entirely absent in existing benchmarks, we include a \novel. We crowdsource 20 novel queries from $N=9$ NLP practitioners (e.g., ``creating Matplotlib visualizations'') to construct this set.
\paragraph{Known Validation.}
To assess if retrieved \samples correlate with performance on established benchmarks, we construct a set of \usecases where test sets specifically designed for that \usecase already exist. For example, we pair the query ``Using meal descriptions to measure calorie intake'' with NutriBench, allowing us to validate if \methodname retrieves items that predict ground-truth performance.  We call these \textit{gold} test sets (see Appendix~\ref{promptwithgoldtestsets} for the mapping from use-cases to tasks). 

\paragraph{Automatic evaluation metrics.}
We grade relevance with \texttt{gpt-5-mini-2025-08-07}, labeling retrieved items \relevant, \partially, or \irrelevant.
When computing metrics, we count both \relevant and \partially items as positive because the goal of \methodname is to surface broadly relevant \samples for multiple plausible interpretations of the query rather than enforcing a single definition of relevance. We report two precision metrics to separate retrieval quality from the effects of filtering: (a) \methodp, which evaluates the unfiltered \samples returned by each anchor method, and (b) \systemp, which evaluates the final \samples returned by the full system. \systemp is intended to track practitioner utility and hence, utilizes a stricter judge prompt (Appendix~\ref{app:evaluation_prompt}), while \methodp is used only to select the anchor method for the final system's design. Additionally, we report ranking quality of the final retrieved set using \ndcg. Finally, we also measure \recall for the \topic; to avoid underestimating recall from relevant \samples outside the designated gold set, we use a restricted 28k-\sample \database containing only gold test sets for \usecases in the \topic. For all other sets, all metrics are reported over a database of 70k-\samples (listed in Appendix~\ref{app:benchmarks}).
% Full metric definitions and judge prompts are in Appendix~\ref{app:eval-metrics}.

% We report two precision variants to isolate the impact of our filtering stage: (a) \methodp:  to compare the quality of the retrieved \samples by different anchor methods before any filtering and (b) \systemp: to compare the overall quality of the system's retrieved \samples. \systemp is the precision that is intended to the correlate with practitioner utility and \methodp is only computed to aid our selection of anchor method for the system's design. To proxy real-world practitioner standards, we use a stricter prompt for \systemp (see \S\ref{app:evaluation_prompt}), a design choice we validate via human evaluation. We also measure \methodname's recall for the \topic where gold test sets for the \usecases already exist in our \database. Only for \topics: we use a smaller \database of approximately 28K \samples which only contains all the gold test sets within the \topics. We do this to avoid supressing \recall over the larger \database which inevitably contained relevant \samples overlapping with the \usecases in the \topics.
% Finally, we measure the final set of retrieved \samples for their ranking quality using \ndcg. Full metric definitions and LLM judge prompts are in Appendix~\ref{app:eval-metrics}.

% Define professional colors for indicators
\definecolor{bestcolor}{HTML}{C6EFCE}      % Soft green for best
\definecolor{secondcolor}{HTML}{FFEB9C}    % Soft gold for second-best

% Helper commands for colored cells
\newcommand{\best}[1]{\cellcolor{bestcolor}\textbf{#1}}
\newcommand{\secondp}[1]{\cellcolor{secondcolor}#1}

\begin{table*}[t]
\centering
\footnotesize
\setlength{\tabcolsep}{2pt}
\caption{Unified \methodname retrieval performance with system and anchor generation evaluation metrics at $k=20$. \testcase and \shorthand have high \systemp. All methods have the least precision on the \textit{applications} set, potentially due to the contextual nature of its queries. Best values per column are highlighted in \colorbox{bestcolor}{\textbf{green}}, with the second-best in \colorbox{secondcolor}{gold}. The horizontal line separates neural embedding-based methods (lower) from non-neural methods (upper).}
\label{tab:unified_metrics}
\begin{adjustbox}{max width=\textwidth}
\begin{tabular}{lcccccccccc}
\toprule
 & \multicolumn{4}{c}{\textbf{Topics}} & \multicolumn{3}{c}{\textbf{Skills}} & \multicolumn{3}{c}{\textbf{Applications}} \\
\cmidrule(lr){2-5} \cmidrule(lr){6-8} \cmidrule(lr){9-11}
\textbf{Method} & \systemp & \methodp & \ndcg & \recall & \systemp & \methodp & \ndcg & \systemp & \methodp & \ndcg \\
\midrule
Random & 0.304 & 0.478 & 0.702 & 0.029 & 0.310 & 0.396 & 0.628 & 0.045 & 0.051 & 0.231 \\
BM25 & 0.529 & 0.748 & 0.850 & 0.191 & 0.654 & 0.780 & 0.823 & 0.320 & 0.309 & 0.648 \\ 
\midrule
\vanilla & 0.763 & 0.887 & 0.921 & 0.350 & 0.768 & 0.923 & 0.911 & \best{0.572} & \best{0.722} & \best{0.870} \\
\segment & 0.812 & 0.896 & 0.923 & 0.345 & 0.775 & \best{0.933} & 0.908 & 0.563 & \secondp{0.687} & 0.849 \\
\shorthand & \secondp{0.884} & \best{0.970} & \secondp{0.955} & \secondp{0.414} & \secondp{0.815} & 0.925 & \secondp{0.916} & \best{0.572} & 0.611 & 0.808 \\
\testcase & \best{0.890} & \secondp{0.968} & \best{0.971} & \best{0.423} & \best{0.837} & \secondp{0.926} & \best{0.937} & 0.560 & \secondp{0.687} & \secondp{0.869} \\
\bottomrule
\end{tabular}
\end{adjustbox}
\end{table*}

\section{\methodname Finds Relevant Examples}
\label{sec:retrieval-eval}
Table~\ref{tab:unified_metrics} compares how well different anchoring strategies retrieve relevant examples.
Across both \topic and \skill, rewriting-based anchors (\segment, \shorthand, \testcase) substantially outperform non-neural baselines (Random, BM25) and the direct \vanilla baseline, with gains of more than 10 points in \methodp and \systemp.
\testcase is consistently the best or second-best method on all three sets, and \shorthand performs comparably. %indicating that rephrasing practitioner \usecases into benchmark-like descriptions is a strong, generally applicable anchoring strategy.
% These gains are not limited to top-$k$ hits:
\ndcg also improves, showing that rewriting improves the entire ranking of retrieved \samples, not just the first few items.
\recall is conservative for all methods on the \topic, which we attribute to
%We attribute the conservative \recall---despite high precision on the \topic---to
the presence of multiple highly overlapping \usecases\ in this set (e.g., high-school vs.\ college chemistry).
Such overlap hurts recall for a large proportion of \usecases because \samples\ relevant to one \usecase\ are counted as misses for a highly-related \usecase.

%\swabha{how exactly are these tie breaking experiments done, a quick explainer might help those not familiar with IR literature}
% addressed in text now! 
\shorthand and \testcase achieve the highest precision across diverse \usecase sets in Table~\ref{tab:unified_metrics}.
We conduct two tie-breaking experiments:
% In addition to achieving the highest precision across diverse \usecases\ on our automatic evaluations (Table~\ref{tab:unified_metrics}), we conduct 2 tie-breaking experiments between \shorthand and \testcase.
(i) we compare how well model performances on retrieved \samples preserve the ranking of models on the gold test sets for the \topic (Appendix~\ref{app:tie-breaking-correlation}).
(ii) For \textit{application} and \skill use-cases which do not have gold test sets, we measure the extent to which
%we use \textit{intersection}@$k$ -
each method's retrieved \samples overlap with the union of all \samples judged relevant by any anchoring method (Appendix~\ref{app:retrieved-set-overlap}).
\testcase\ performs best in both experiments;
% Across both experiments, \testcase\ yields the strongest performance correlations with known gold test sets and the highest approximate recall (\S\ref{app:tie-breaking}, \S\ref{app:retrieved-set-overlap});
we therefore use \testcase\ as the anchor generation method in all subsequent experiments. 

We also study \systemp for \testcase as we vary $k$ (Appendix~\ref{app:precision-as-k-increases}). It is largely stable for the \topic and \skill as $k$ grows, suggesting \methodname\ can return larger pools of \samples without rapidly flooding practitioners with irrelevant items.
The \application observes a sharper degradation in \systemp as $k$ grows potentially due to the paucity of \samples that are relevant to the highly contextual, nuanced \usecases in this set.

To also validate these judgments under another indepenent LLM judge, we further replicate these evaluations with \texttt{gemini-2.5-flash-lite} in Appendix~\ref{app:automatic-judge-calibration} and observe strong, statistically significant correlations ($\rho=0.795$ for \systemp) against our judge (Table~\ref{tab:spearman_correlations} in the appendix).

\paragraph{\methodname is fast.}
We report end-to-end user-perceived latency for the public \methodname website over 210 requests (0 failures). Latency of the end-to-end warm-started system is $5.28\pm1.48$ s (mean and  stdev).
Detailed distribution statistics (tail, median and per-stage costs) and hosting infrastructure are reported in Appendix~\ref{app:infrastructure}.
\methodname enables iterative, diverse audits of practitioner \usecases at interactive speeds.

\subsection{Human Evaluation of \methodname}
\label{sec:human-eval}

We recruit $N=10$ NLP practitioners to annotate retrieval relevance using the same three-point scale as our LLM judge. For all the \usecases\ we consider, every \sample in the top-20 \samples retrieved by \methodname is annotated by three annotators. 
To rigorously audit the judge, we employ two sampling strategies to select the \usecases for annotation.  
First, we ask annotators to annotate \samples for \underline{all} \usecases in the \novel and \known to establish a baseline of correlation between automatic and human judges. 
Through this we collect $n=\num{2460}$ relevance judgments. 
Second, for the \topic, \skill and \application, we take the 10 \usecases with the highest precision under the LLM judge and have their relevance judgments verified by the annotators.
This setup allows us to determine if an automatic judge's high scores reflect genuine relevance or merely the systematic leniency often observed in LLM-based evaluation \citep{jain2025beyond, arabzadeh2025benchmarking}. 
Through this we collect $n=\num{1800}$ relevance judgments. In all, we collect $n=\num{4260}$ judgments across 71 \usecases.
Full details are in Appendix~\ref{app:humanevaluation}.

\paragraph{Alignment between automatic and human judgments.}
We calculate mean relevance judgments by counting \relevant as 1, \partially as 0.5, and \irrelevant as 0.
Table~\ref{tab:judge-validation-summary} compares human relevance judgments with automatic \systemp\ via paired $t$-tests on mean relevance (bias) and Spearman’s $\rho$ (rank agreement). For the \textit{topics}, \textit{skills}, and \textit{applications} sets, mean relevance is closely matched (mean difference $\le$0.1), indicating that higher automatic \systemp\ generally tracks higher human precision.
%rather than systematically inflating it.
Rank agreement is weak-to-moderate across sets ($\rho\in[0.15,0.50]$). The \novel\ is moderately aligned (mean 0.67 vs.\ 0.70). The \known\ is the main outlier (mean gap 0.30); error analysis traces this to a few \usecases with simplified queries too broad for the esoteric suites they target (Appendix~\ref{app:judge-error-analysis}). Overall, automatic judgments provide a reasonably calibrated proxy for practitioner-assessed precision, while occasionally misestimating absolute relevance for broad, domain-specific \usecases.

% Rewrote this to focus on how the novel set (which is also all-usecase evaluation rather than top-10 shows good/in fact pessimistic automatic)
% Table~\ref{tab:judge-validation-summary} compares human and automated \systemp. We compare human and automatic scores via (i) paired $t$-tests on mean relevance (bias) and (ii) Spearman’s $\rho$ (rank agreement). For four of five sets (\novel, \topic, \skill, \application), mean differences are $\le 0.10$ and correlations are positive ($\rho\in[0.15,0.50]$; Table~\ref{tab:judge-validation-summary}), suggesting that higher automatic \systemp\ corresponds to higher human precision (i.e., the judge is not over-optimistic in its assessment). The \known\ is an outlier (mean gap $0.30$); upon investigation we find that the gap for this set is driven by $5$ \usecases whose simplified \usecase queries appear to be too broad for the esoteric test sets for which these \usecases were designed (see Appendix~\S\ref{app:judge-error-analysis} for examples).
% Overall, the  weak-to-moderate correlations across all sets suggest that automatic judgments provide a reasonably calibrated proxy for practitioner-assessed precision.

\paragraph{Inter-annotator agreement (IAA) and subjectivity.}
IAA for the relevance judgments ranges from slight to moderate (Table~\ref{tab:judge-validation-summary}). Error analysis finds that disagreements are between \relevant\ vs.\ \partially\ rather than \relevant\ vs.\ \irrelevant\ (Figure~\ref{fig:label-distribution-comparison} in Appendix~\ref{app:judge-error-analysis}). This pattern suggests calibration differences rather than fundamental disagreement, consistent with prior work on the subjectivity of relevance labels \citep{adams2023sparse,DBLP:conf/hcomp/McDonnellLKE16,shirani2021learning, el2020investigating}.

% Inter-annotator agreement (IAA) for the collected relevance judgments is mixed: Fleiss’s~$\kappa$ is moderate for the \application and \known ($\kappa\,{\approx}\,0.29$) but slight for the remaining sets (Table~\ref{tab:judge-validation-summary}). This range of correlation is consistent with prior work eliciting subjective labels such as ``relevance,'' where differences largely reflect calibration rather than fundamental disagreement \citep{adams2023sparse,DBLP:conf/hcomp/McDonnellLKE16,shirani2021learning, el2020investigating}.
% We confirm that even for the annotations we elicit, the maximum  disagreements are concentrated in the \relevant\ vs.\ \partially\ boundary rather than \relevant\ vs.\ \irrelevant\ inversions (Table~\ref{tab:judge-validation-summary}, \S\ref{app:judge-error-analysis}).  
\begin{figure*}[t!]
    \centering
    \includegraphics[width=0.9\linewidth]{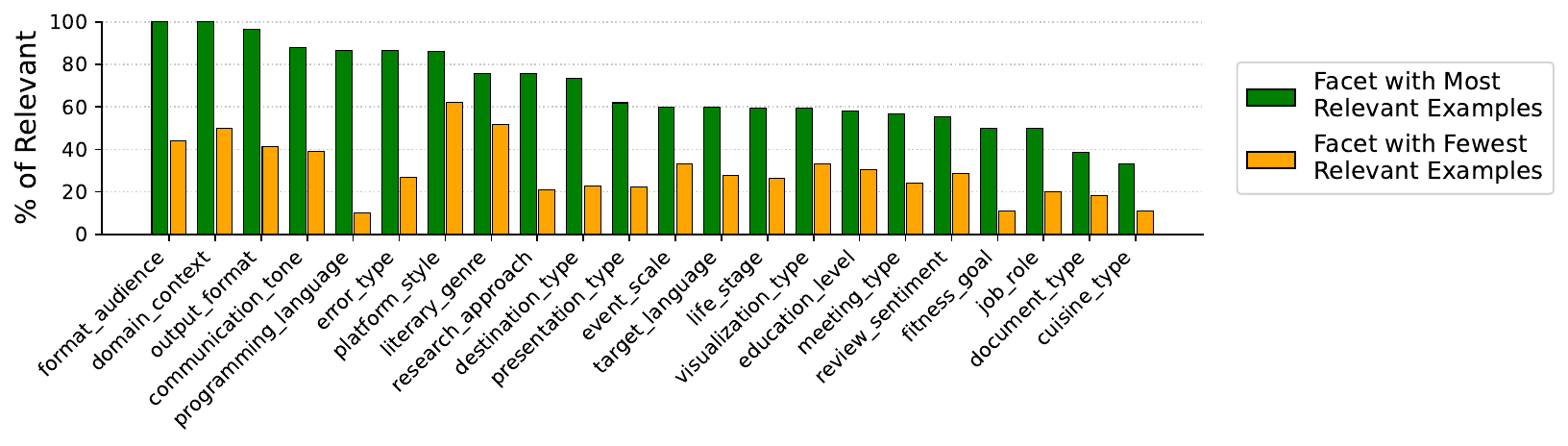}
    \caption{Variation in the percentage of \relevant \samples across different facets of the same broad \usecase. For each skill family, we show the facet for which the highest and lowest number of \samples are retrieved. Most capabilities show sharp disparities: certain facets are heavily over-represented in our database of 70k benchmark \samples, while others are nearly absent from benchmarks.}
    \label{fig:content_validity}
\end{figure*}

\section{Measuring Content Validity}
\label{sec:content-validity}
Having established that \methodname retrieves relevant \samples, we next demonstrate how \methodname can help practitioners collect evidence for measuring  
\textit{content validity}, the extent to which an evaluation comprehensively represents all facets of the ability it claims to measure \citep{Strauss2009ConstructVA,freiesleben2025benchmarking,wallach2025position}. In our setting, this means asking whether \samples across benchmarks adequately cover both the breadth (different facets) and depth (difficulty or context) of capabilities that practitioners care about. Establishing the content validity of an assessment is crucial. If benchmarks collectively instantiate a capability narrowly, important facets can be left untested, and aggregate scores can systematically distort a model’s competence. If \samples for ``code generation’’ mostly involve short Python snippets, then performance on those \samples may tell us little about how a model fares on code generation for other languages.

For a practitioner, one way to assess content validity is to evaluate a capability across different facets: hold the broad capability fixed (e.g., ``code generation'') while varying a single facet (e.g., programming language) and checking benchmark coverage \cite{landsheer2010search}. \methodname\ makes coverage gaps directly inspectable by retrieving the relevant \samples\ for each facet.
% We use \methodname to probe how well existing benchmarks cover different facets of various realistic \usecases. Concretely, we ask: if we hold a broad capability fixed (e.g., ``code generation’’ or ``debugging code’’) and vary a single facet (e.g., programming language or error type), do current benchmarks represent those variants comparably well, or are some facets absent?

% %\swabha{nit: "testset" is usually not used as a single word, it might be preferable to use "test sets".} 
% % Starting from the \textit{Skills} query set introduced in \S\ref{evaluation-categories},   
% To answer this question, we need to compare the retrieved \samples for \usecases that focus on the same broad capability and only vary across very specific facets.
 
\paragraph{Experimental setup.}
To evaluate \methodname\ in this setting, we need \usecases\ paired with relevant facets and explicit axes of variation.
%Ideally, these facets would be elicited directly from practitioners in situ; but for the purpose of this demonstration,
We use the \skill to seed \texttt{\small claude-sonnet-4-5-20250929} to produce 20 broad capabilities and six closely related variants for each, differing along a single, interpretable axis (120 total). For instance, ``trip planning'' varies by destination type (e.g., domestic vs.\ international). We manually verify that variants are lexically separable,
% (e.g., ``city'' vs.\ ``village''),
ensuring that a purely lexical baseline could, in principle, distinguish them. We call the variants of a capability a \emph{skill family}. Table~\ref{tab:content-validity-examples} gives two example skill families.
%(code generation, creative writing).

\vspace{-8pt}

\paragraph{Evaluation.}
Estimating content validity in practice is expensive because both (i) which facets matter and (ii) what counts as ``relevant'' are stakeholder-dependent \citep{Strauss2009ConstructVA}. Moreover, variants within a fixed skill can often retrieve overlapping \samples (e.g., items relevant to ``Java programming'' may be partially relevant to ``Python programming''), forcing annotators to repeatedly judge near-duplicates across facets---an especially high cognitive burden \cite{10.1145/3290605.3300522, henley2024supporting}. Accordingly, we use the LLM judge
% that we verified in the previous section
as a proxy for relevance judgments.
We compute the percentage of \samples retrieved by \methodname deemed \relevant by the LLM judge (excluding \partially \samples).
Large differences between a skill family's variants signals low content validity \cite{landsheer2010search}.

\vspace{-8pt}
%  \swabha{do u mean table 5?}

\definecolor{sigblue}{RGB}{0, 90, 156}
\definecolor{sigorange}{RGB}{213, 94, 0}
\newcommand{\siga}{\textsuperscript{\textcolor{sigorange}{$\dagger$}}}
\newcommand{\sigb}{\textsuperscript{\textcolor{sigblue}{$\ddagger$}}}

\begin{table}[t]
\centering
\footnotesize
\caption{Mean relevance of retrieved \samples by human annotators vs. the LLM judge across use-case query sets. Superscripts indicate significance: \siga~$p<0.05$; \sigb~$p<0.0001$. Markers on Mean\textsubscript{LLM} are significance markers for paired $t$-test for the means. $\rho$ is Spearman's rho and $\kappa$ is Fleiss's kappa.}
\label{tab:judge-validation-summary}
\begin{tabular}{@{} l S[table-format=1.2] S[table-format=1.2] S[table-format=1.2] S[table-format=1.2] @{}}
\toprule
\textbf{Use-case} & {\textbf{Mean\textsubscript{H}}} & {\textbf{Mean\textsubscript{LLM}}} & {${\rho}$} & {${\kappa}$} \\
\midrule
Topics           & 0.96 & 0.92\sigb & 0.15\siga & 0.10 \\
Skills           & 0.80 & 0.91\sigb & 0.17\siga & 0.11 \\
Applications     & 0.74 & 0.75      & 0.50\sigb & 0.29 \\
Known Valid.     & 0.58 & 0.88\sigb & 0.34\sigb & 0.29 \\
Novel            & 0.70 & 0.67      & 0.27\sigb & 0.13 \\
\bottomrule
\end{tabular}
\end{table}
% \definecolor{sigblue}{RGB}{0, 90, 156}
% \definecolor{sigorange}{RGB}{213, 94, 0}

% \newcommand{\siga}{\textsuperscript{\textcolor{sigorange}{$\dagger$}}}
% \newcommand{\sigb}{\textsuperscript{\textcolor{sigblue}{$\ddagger$}}}

% \begin{table*}[htbp]
% \centering
% \small
% \caption{
% \greg{Make a succinct one-column version and move this full version to the appendix.}
% The mean relevance of retrieved \samples as determined by human annotators is very similar to that of the LLM relevance judgments for four out of five use-case query sets.
% $\Delta$: difference in mean relevance between human annotators and the LLM judge; \siga~significant at $p<0.05$;
% \sigb~significant at $p<0.0001$.}
% \begin{adjustbox}{max width=\textwidth}
% \begin{tabular}{@{} l S[table-format=1.2] S[table-format=1.2] S[table-format=+1.2] S[table-format=+2.2] S[table-format=1.2] S[table-format=1.2] @{}}
% \toprule
% \textbf{Use-case query set} & {\textbf{Mean\textsubscript{human}}} & {\textbf{Mean\textsubscript{LLM}}} & {\textbf{$\Delta$}} & \textbf{Paired $t$-test} & \textbf{Spearman's $\rho$} & \textbf{Fleiss's $\kappa$} \\
% \midrule
% Topics           & 0.96 & 0.92 & -0.04 & -2.98\sigb  & 0.15\siga  & 0.10 \\
% Skills           & 0.80 & 0.91 & +0.10 & 5.40\sigb   & 0.17\siga  & 0.11 \\
% Applications     & 0.74 & 0.75 & +0.01 & 0.31       & 0.50\sigb  & 0.29 \\
% Known Validation & 0.58 & 0.88 & +0.30 & 17.70\sigb  & 0.34\sigb  & 0.29 \\
% Novel            & 0.70 & 0.67 & -0.03 & 1.26        & 0.27\sigb  & 0.13 \\
% \bottomrule
% \end{tabular}
% \label{tab:judge-validation-summary}
% \end{adjustbox}
% \end{table*}

\paragraph{Findings.}
% \swabha{this figure is a little confusing to me, what do individual bars represent? why do some bars not add to a 100?}
% Changed to the bar plot. 
Figure~\ref{fig:content_validity} summarizes the percentage of \relevant \samples for the best- and worst- represented facet of each skill family.
Closely related \usecase variations yield very different numbers of \relevant \samples, indicating sizable spread in facet-level coverage in each skill family. Table \ref{tab:content-validity-examples} gives examples of this spread: dominant formulations such as ``write a Python function’’ or
%``debug a syntax error’’
``write a fantasy story''
are associated with high yields of \relevant \samples, whereas equally legitimate but less conventional facets such as ``write code in Go’’ or
%``debug a memory leak’’
``write a historical fiction story''
often drop precipitously, despite testing the same underlying capability. 
Appendix~\ref{app:content-validity} finds the same trends with a lexical BM25 baseline, suggesting that our findings are due to benchmark composition rather than an artifact of the semantic retriever.

\begin{table}[t!]
\centering
\footnotesize
\setlength{\tabcolsep}{3pt}
\renewcommand{\arraystretch}{1.0}
\caption{
Different facets of the same skill family can have very different \systemp when using \methodname with Example Synthesis anchors and the LLM judge (relevant \samples only). Most skill families show substantial spread: some variants (green) have many \relevant \samples, while others (red) are barely represented. For complete results for all skill families, see Appendix~\ref{app:content-validity}.}
\label{tab:content-validity-examples}
\vspace{-3pt}
\resizebox{\linewidth}{!}{%
\begin{tabular}{p{7em}>{\centering\arraybackslash}p{5em}p{7em}>{\centering\arraybackslash}p{5em}}
\toprule
\textbf{Code generation} & & \textbf{Creative writing} \\
\multicolumn{2}{p{12em}}{\textbf{Programming language}} & \multicolumn{2}{p{12em}}{\textbf{Literary genre}} \\
\multicolumn{2}{l}{``Write a function in $\ldots$''} & \multicolumn{2}{l}{``Write a $\ldots$ story''} \\
\midrule
Facet & \systemp & Facet & \systemp \\
\midrule
\cellcolor{green!20}Python & \cellcolor{green!20}0.867 & \cellcolor{green!20}fantasy & \cellcolor{green!20}0.759 \\
C++ & 0.741 & science fiction & 0.679 \\
Rust & 0.478 & mystery & 0.654 \\
JavaScript & 0.346 & romance & 0.625 \\
Java & 0.286 & horror & 0.583 \\
\cellcolor{red!20}Go & \cellcolor{red!20}0.100 & \cellcolor{red!20}historical fiction & \cellcolor{red!20}0.517 \\
\bottomrule
\end{tabular}
}
\end{table}

The disparities uncovered here have direct consequences for how practitioners interpret benchmark scores; 
for example, if a benchmark reports a single number for debugging code and almost all retrieved \samples concentrate on a single facet (e.g. syntax errors), then strong performance may say little about the model’s behavior on other deployment-critical variants (e.g., security vulnerability or memory leaks). \methodname can help practitioners uncover such disparities and can ground the construction of new benchmarks across under-explored facets of that skill. 

\section{Measuring Convergent Validity}
\label{sec:convergent-validity}

With the rapid proliferation of benchmarks, it is now common to find multiple benchmarks that ostensibly evaluate the same capability \cite{sun2025survey, zhang2025redundancy}. 
This invites an implicit but often untested assumption: if two benchmarks truly measure the same underlying skill, they should lead to similar conclusions about which models are better for that skill. \textit{Convergent validity} formalizes this expectation: model rankings on benchmarks that purport to test the same skill should be correlated~\citep{Strauss2009ConstructVA, Bean2025MeasuringWM}. 
When this assumption fails, benchmark results become hard to trust or act on, yielding contradictory takeaways
%such as \textit{Model A beats Model B} on one benchmark but reverses on another~
\citep{diddee2024chasing, zhou2025lost}.
\methodname helps practitioners diagnose these failures of convergence by retrieving relevant \samples that enable the construction of alternate, \usecase specific test sets. 
Practitioners can evaluate models on these alternate test sets and check whether specialized benchmark-derived rankings are preserved on them. As we demonstrate in this section, this exercise can reveal when “same capability” benchmarks do (or \textit{do not}) support transferable conclusions.

\paragraph{Experimental setup.}
We use the \known of \usecases (\S\ref{sec:evaluation-categories}). For each \usecase $u$, we evaluate $n=8$ models on (i) its gold test set $G_u$ and (ii) the retrieved set $R_u$. For each set, we score \samples using the source-benchmark metric, aggregate to model-level scores, and induce a ranking.
Model and evaluation details are in Appendix~\ref{app:convergent-validity}.
For each benchmark, we compute the per-sample ``success'' of the model by using its designated success scoring metric (win rate, accuracy, exact match, etc.)
and aggregate the per-sample ``success'' under the designated benchmark to compute the models' performance across the retrieved \samples
(see Appendix~\ref{app:benchmark-metrics} for detailed metric computation).
Then, we measure rank agreement via two forms of Kendall's $\tau$ (using the mean across 50 Monte Carlo trials):
\begin{enumerate}
    \item $\tau_{\text{ret}}(u)$: shows the agreement between the ranking induced by $R_u$ and a gold ranking estimated at matched cardinality. In each trial, for each model $m$, we subsample $|R_{u}|$ examples from $G_u$, induce a ranking, and report its mean Kendall's $\tau$ with the ranking from $R_u$.
    \item $\tau_{\text{gold}}(u)$: a baseline of the correlation between model rankings on the full gold set with model rankings on a smaller subset of that gold set. Let $k$ be the (average) retrieved size for $u$; in each trial we subsample $k$ examples from $G_u$, induce a ranking, and compute Kendall's $\tau$ against the ranking on the full gold set. This accounts for the decrease in the maximum achievable Kendall's $\tau$ for smaller sets \citep{yauney2025reliable}.\footnote{Appendix~\ref{app:convergent-validity} reports correlations between same-cardinality subsets as a sanity check.}
\end{enumerate}

We define \emph{rank divergence} as $\Delta(u)=\tau_{\text{gold}}(u)-\tau_{\text{ret}}(u)$. Small $\Delta$ indicates retrieved \samples preserve rankings, while large $\Delta$ reflects divergence beyond finite-sample effects, indicative of low convergent validity. 

\paragraph{Evaluation.}
Because convergence claims require that retrieved sets reflect the intended \usecase, we strictly restrict analysis to \usecases whose retrieved \samples are verified as relevant during our human evaluation. Concretely, we report results only for use-cases where the mean human relevance over \samples is at least \partially.

\begin{figure}[t!]
\centering 
\vspace{10pt}
\includegraphics[width=0.9\linewidth]{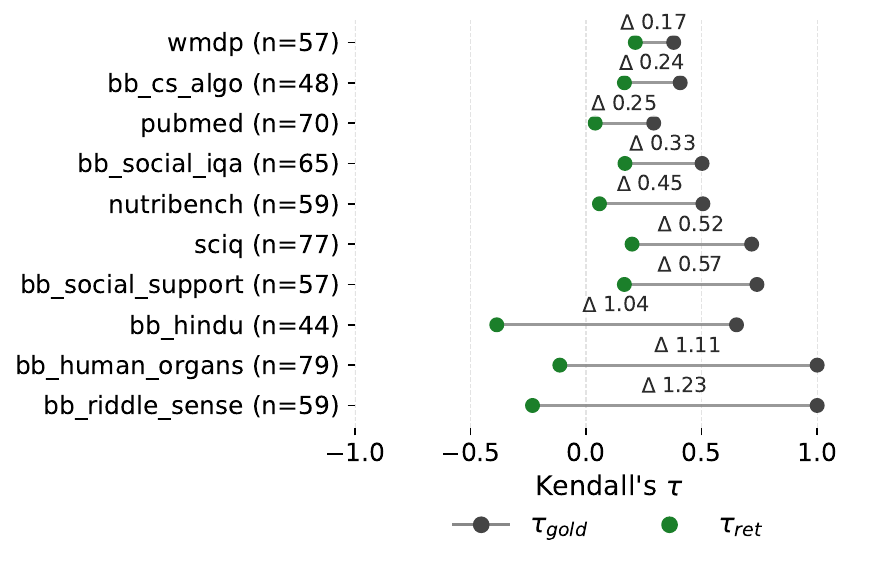} 
\caption{ Kendall’s $\tau$ correlations of model rankings on gold test sets vs.\ retrieved sets for \known \usecases with mean human relevance $\geq 0.5$. Wider bars correspond to greater rank divergence $\Delta = \tau_\text{gold} - \tau_\text{ret}$. $n$: number of retrieved examples. } 
\label{fig:relevance-vs-tau} 
\end{figure}

\paragraph{Findings.}
Figure~\ref{fig:relevance-vs-tau} shows rank divergence varies sharply across \usecases\; Some exhibit low divergence, indicating that the retrieved sets preserve benchmark-induced model comparisons.
Others show large divergence, even including ranking inversions, suggesting that high topical relevance may not be sufficient for establishing convergent validity.

Manual inspection of the retrieved \samples suggests three recurring drivers of ranking inversions (Table~\ref{tab:ranking-inversions}). Collectively, they reflect two practical ways in which practitioners are likely to encounter low convergent validity. 

\definecolor{nude}{HTML}{228B22}
\definecolor{gold}{HTML}{D4AF37}
\definecolor{lightblue}{HTML}{B0D4E8}

\begin{table}[t!]
\centering
\small
\setlength{\tabcolsep}{4pt}
\caption{Root causes of ranking inversions in \methodname; 
\colorbox{gold!30}{gold} highlight indicates \sample is from the gold set, and \colorbox{lightblue!30}{blue} indicates it is from the retrieved set.}
\label{tab:ranking-inversions}
\resizebox{\linewidth}{!}{%
\begin{tabular}{@{}p{0.95\columnwidth}@{}}
\toprule
\textbf{Root Cause 1: Rubric Shift} \\
\textit{\textbf{Example Query:} ``Hindu Religion''} \\
\cellcolor{gold!30}\textbf{HellaSwag:} Philosophy and Religion: How to learn about ancient hinduism ... \\
\cellcolor{lightblue!30}\textbf{BIG-bench Hard (Hindu Knowledge):} Which of the following gods are associated with a flying elephant \\
\midrule
\textbf{Root Cause 2: Divergent Operationalization} \\
\textit{\textbf{Example Query:} ``Solving Puzzles''} \\
\cellcolor{gold!30}\textbf{BIG-bench Hard (Colored Objects):} On the desk, you see items in a row: a gold textbook, a purple puzzle, a teal necklace, and a silver pencil. How many non-gold items do you see to the right of the pencil? \\
\cellcolor{lightblue!30}\textbf{AlpacaEval:} Hi, I'm trying to solve a crossword puzzle, but I've never done one of these before. Can you help me out? \\
\midrule
\textbf{Root Cause 3: Context Shift (Recall → Applied)} \\
\textit{\textbf{Example Query:} ``Writing Programs''} \\
\cellcolor{gold!30}\textbf{HumanEval:} Check if in given list of numbers, are any two numbers closer to each other than given threshold. \\
\cellcolor{lightblue!30}\textbf{InfoBench:} ... you've came up with an algorithm for solving the coding question your interviewer gave you but ... \\
\bottomrule
\end{tabular}
}
\vspace{-5pt}
\end{table}

First, retrieved \samples\ often vary the \emph{format or scoring rubric} of the evaluation (e.g., short-form factual questions vs.\ longer, constraint-heavy, or open-ended prompts). Hence, even when an example's topic matches the \usecase, such shifts change what it means to succeed (Root Cause~1 in Table~\ref{tab:ranking-inversions}). This creates a measurement mismatch: different metric and rubrics can systematically alter observed model rankings for the same nominal capability due to inherent differences in what the measurements reward \cite{campbell1959convergent}. For instance, win-rate is a popular metric for generative tasks; it is known to be sensitive to calibration inconsistencies \cite{dubois2024length, gao-etal-2024-bayesian, zheng2024cheating}, which can propagate into rank differences and ultimately yield divergent conclusions.

Second, for over-represented capabilities with dedicated benchmarks, retrieval may concentrate on one source whose \emph{operationalization} of the advertised skill is narrower or differently scoped (e.g., two ``puzzle-solving'' benchmarks emphasizing distinct task families). In this case, benchmarks that appear to target the same capability may in fact measure different underlying competencies, yielding \emph{construct non-equivalence} (Root Cause~2 in Table~\ref{tab:ranking-inversions}). Scoping differences can also appear as a shift from definitional recall to applying a concept in context, effectively redefining the construct under evaluation (Root Cause~3 in Table~\ref{tab:ranking-inversions}).

\methodname\ helps diagnose validity threats by generating clear evidence for an observed rank instability: when benchmarks expected to measuring the same “capability” produce divergent model rankings, surfacing the retrieved \samples\ and their source benchmarks can help practitioners manually inspect whether the inconsistency reflects construct non-equivalence or divergent operationalization.

\section{Related Work}
\paragraph{Customizable benchmark selection and construction.}
Customizable benchmarking also addresses the sparse coverage and narrow operationalizations of existing benchmarks.
These approaches synthetically generate test sets that tailor evaluation to a practitioner’s task or domain from a seed corpus \citep{shashidhar2025yourbench} or in a zero-shot manner \citep{pombal2025zero}.
%, e.g., \citet{shashidhar2025yourbench} (corpus-conditioned test generation) and \citet{pombal2025zero} (fully synthetic, zero-shot benchmarks).
Others re-structure existing evaluations to enable targeted subset selection, e.g., capability meta-descriptions \citep{zeng2025evaltree}, description-driven retrieval of difficult items \citep{li2024autobencher}, and large benchmark repositories with fixed taxonomies \citep{kim2025benchhub}. In contrast, we retrieve and organize \emph{vetted, labeled} benchmark items for a practitioner’s free-form intent, avoiding both synthetic labels and reliance on predefined schemas.
This preserves the manual supervision that was applied to the test set curation originally, reducing sensitivity to artifacts from LLM-generated evaluations \citep{huang2024bias, chen2024llms}.
This choice aligns with evidence-centered benchmark design, which posits that capability claims should be grounded in item-level evidence \citep{liu-etal-2024-ecbd}. 

\paragraph{Infrastructure for validity-oriented evaluation.}
Benchmark audits emphasize that modern evaluation practice often under-specifies validity \citep{dietz2025llm, reuel2024betterbench}. Yet establishing validity typically requires construct-sensitive measurement tools \citep{salaudeen2025measurement, ye2025large} and expensive evidence collection: expert annotation \citep{Bean2025MeasuringWM}, explicit enumeration of operationalizations \citep{wallach2025position}, and stakeholder-aware construct definitions \citep{alaa2025medical}. While psychometrics-inspired approaches, such as IRT-based fluid benchmarking \citep{hofmann2025fluid}, improve test composition by modeling item properties, practitioners lack scalable support for assembling and inspecting the item-level evidence needed to surface missing coverage and conflicting operationalizations, a gap our retrieval-based interface fills.

\section{Discussion and Conclusion}
We present \methodname, a meta-evaluation tool for \emph{cost-effective, transparent} validity assessment of LLM benchmarks under practitioner-defined \usecases. Rather than treating benchmark scores as self-explanatory, \methodname\ surfaces the underlying items that instantiate a \usecase, enabling practitioners to directly audit \emph{what} is being measured and \emph{how}. Human evaluations, consistent with LLM-judge results, show that \methodname\ retrieves relevant evidence for assessing coverage and representation. By making sources, rubrics, and framing differences inspectable, \methodname\ supports rapid diagnosis of low content validity and convergent validity.

\methodname helps audit the validity of LLM evaluations by surfacing evidence \emph{across} benchmarks rather than relying on any single suite \cite{veuthey2025meqa, son2024mm, Chern2024CanLL}. Because capabilities admit multiple operationalizations, this cross-benchmark view makes it practical to inspect the most atomic evidence for a \usecase—the evaluation items themselves—revealing which subsets, outliers, or framings drive ranking inversions and where facets are underrepresented. In doing so, \methodname supports more stable, well-contextualized conclusions and directly guides where new benchmarks (or slices of existing ones) should focus to better measure the intended capability.

\section*{Impact Statement}
Systems are only as good as the yardsticks we measure them with. We take a first step toward practitioner-centered validation by offering a fast, low-cost, end-to-end workflow for auditing benchmark validity under user-specified intent. Rather than treating disagreement across benchmarks as noise, we expose the underlying measurement choices---rubrics, framing, and construct scope---that can produce conflicting rankings, allowing practitioners to interpret competence claims more cautiously and to diagnose where (and why) conclusions diverge.

Our work contributes a core ingredient for trustworthy benchmarking by making capability claims auditable \cite{cheng2025benchmarking}. When user-centric capabilities---rather than metric dominance on traditional tasks---drive utility, practitioners need direct access to the evaluation items and rubrics that underwrite system-selection decisions.
We do not anticipate any negative ethical impacts of this work.

\section*{Acknowledgments}
This research was supported by a Gemini Credits Grant. We thank Fernando Diaz, whose course first introduced us to the concept of validity and its central role in measurement. We are grateful to Joachim Baumann for insightful feedback that helped improve the clarity of this draft. We also thank Kabir Ahuja and Valentin Hofmann for helpful discussions on validity, Varsha Kishore for helpful discussions about retrieval, and members of SS’s and DI’s groups for their comments on improving the clarity of the work. Finally, we thank Saujus Vaduguru and Kyzyl Monteiro for their assistance in organizing the user study for \methodname.

\bibliography{example_paper}
\bibliographystyle{icml2026}

%%%%%%%%%%%%%%%%%%%%%%%%%%%%%%%%%%%%%%%%%%%%%%%%%%%%%%%%%%%%%%%%%%%%%%%%%%%%%%%
%%%%%%%%%%%%%%%%%%%%%%%%%%%%%%%%%%%%%%%%%%%%%%%%%%%%%%%%%%%%%%%%%%%%%%%%%%%%%%%
% APPENDIX
%%%%%%%%%%%%%%%%%%%%%%%%%%%%%%%%%%%%%%%%%%%%%%%%%%%%%%%%%%%%%%%%%%%%%%%%%%%%%%%
%%%%%%%%%%%%%%%%%%%%%%%%%%%%%%%%%%%%%%%%%%%%%%%%%%%%%%%%%%%%%%%%%%%%%%%%%%%%%%%
\newpage
\appendix
\onecolumn

\newtcolorbox{takeaway}{
   breakable,
  colback=lightgray, % Background color of the box
  colframe=black, % Frame color
  boxrule=0.2mm, % Frame thickness
  arc=2mm, % The arc of the box corners
  top=2mm, % Top margin within the box
  bottom=2mm, % Bottom margin within the box
  left=2mm, % Left margin within the box
  right=2mm, % Right margin within the box
  boxsep=0mm, % Space between text and frame in all directions
}
\newcolumntype{L}{>{\raggedright\arraybackslash}X}

\section{Motivation Experiment Details for Figure~\ref{fig:teaser}}
\label{app:motivation-experiment}

We evaluate a set of seven instruction-tuned models, denoted as $\mathcal{M} = \{m_i\}_{i=1}^7$. Specifically, we define $m_1, \dots, m_7$ as \texttt{Qwen2.5-7B-Instruct} \cite{yang2024qwen2}, \texttt{Ministral-8B-Instruct-2410}, \texttt{Llama-3.1-Tulu-3-8B} \cite{lambert2024tulu}, \texttt{c4ai-command-r7b-12-2024} \cite{cohere2025command}, \texttt{DeepSeek-R1-Distill-Llama-8B} \cite{guo2025deepseek}, \texttt{Mistral-7B-Instruct-v0.2} \cite{jiang2023mistral7b}, and \texttt{Meta-Llama-3-8B-Instruct} \cite{grattafiori2024llama}, respectively.

\section{Anchor Generation Methods}
\label{app:anchor-generation}

\subsection{Shorthand Anchor}
\label{app:shorthand_pipeline}
Many benchmarks (e.g., \textsc{Winogrande}, BIG-bench) lack fine-grained capability annotations, making it difficult to identify which skills are being tested at the datapoint level.
To ensure semantic proximity among functionally similar samples across diverse formats, we design structured textual representations called \textit{shorthands}.

\paragraph{Shorthand format.}
A shorthand follows the template 
$\langle\text{skill}\rangle$ \& $\langle\text{key}_1\rangle$ \& $\langle\text{key}_2\rangle$ \& $\langle\text{key}_3\rangle$, 
where $\langle\text{skill}\rangle$ denotes the primary cognitive ability required, and keys encode salient lexical or contextual attributes that enhance retrieval precision.

\paragraph{Pipeline.}
\begin{enumerate}
    \item \textbf{Seed Data construction.} To train a model to efficiently translate our index into the \shorthand representation, we needed to generate a small, high quality translation dataset where a \sample corresponds to its equivalent shorthand. To achieve a diverse set of \samples to act as this set - we first embed 10 random subsets of our entire index using the \texttt{BAAI bge-en-icl embedding model}. Each subset contains 1000 \samples (without replacement). We, then, run t-SNE on each subset separately and k-means per batch (with $k=10$). Finally, we pick, both, the $n=250$ nearest-to-centroid and $n=250$ farthest-from-centroid examples in each cluster. Choosing these allows us to cover both, the prototypical and well represented tasks as well as the least represented tasks for the initial seed data construction. This allows us to generate a dataset consisting of 5000 \samples. 

    \item \textbf{Finetuning Data Synthesis.} We then use \texttt{gpt-5-mini-2025-08-07} to produce shorthands for the 5000 \samples. We use the following prompt for this conversion.

    \begin{takeaway}
    Shorthand Generation Prompt 
    """You are tasked with converting natural language text into a concise shorthand format which can be used to identify other datapoints having similar capabilities.
    The format for creating this shorthand is: $<skill>$ \& $<$key1$>$ \& $<$key2$>$ \& $<$key3$>$ where:
    
    - $<$skill$>$ encodes the primary cognitive/technical ability needed to accomplish the task underlying the text. Ask yourself: "What is the primary cognitive/technical ability needed to accomplish the task underlying the text?"
    Some examples (NOT EXHAUSTIVE): creative\_writing, coding, equation\_solving, coreference\_resolution, strategy\_planning, logical\_reasoning, factual\_recall, reading\_comprehension, scientific\_visualization, graphics, social\_interaction, movie\_trivia, statistical\_analysis, etc. 
    
    - $<$key1$>$,$<$key2$>$, $<$key3$>$ are 1-3 SPECIFIC and DISTINCTIVE features that maximize retrieval precision; For adding each feature/key, ask yourself:
      - "What are the specific document types, topics, and/or entities in this text" that need to be preserved for retrieving similar datapoints?
      - Is there any specific topic underlying being asked by the user query: "machine learning" "statistics" "poetry" etc - include that as a key. 
      - Is there any specific output format mentioned in the query: "research papers", "clinical cases", "financial reports" "emails" etc - include that as a key. 
      - Is the the query about a specific person, organization, or entity - include that as a key. If there are multiple entities, or placeholders like "X", "Y", "Z" - include the broad category of the entity (e.g., "actor", "writer", "sportsperson" "female") and number of entitites as a key.
      - Is the user looking for a specific task like: "sorting\_ascending", "counting\_3\_objects", "implementing\_neural\_network"- include that as a key.  - If the text includes a mathematical formula: include the concept type of the formula (e.g., "differential\_equations", "linear\_algebra", "probability\_distribution")
      - If the user has only specificed a single topic - use that as the key. DO NOT ADD A SKILL UNLESS ITS SPECIFIED. 
      - If the user has only specified a single skill/task - use that skill/task as the key. DO NOT ADD A DIFFERENT TASK UNLESS ITS SPECIFIED. 
      - Use underscores for multi-word concepts (e.g., "machine\_learning", "social\_skills", "new\_york")
      - Make sure that no key adds a new generic domain/skill/topic to the query. STRICTLY FOLLOW THIS RULE.
      - Separate each key with an ampersand.
      - Return exactly 1 shorthand for a user query wrapped in XML tags: $<$shorthand></shorthand$>$. 
    - IMPORTANT: Only include keys, if they add unique, searchable value. If there are no more distinctive features, USE FEWER KEYS rather than padding with generic terms.Your final outputs should be of the form:
        $<$shorthand> </shorthand$>$
      Here are some examples of shorthands:   
    \end{takeaway}
    
    \item \textbf{Model Finetuning.} We then finetune \texttt{Llama-3-8B-Instruct} via on (\sample, \shorthand) pairs.   We fine-tuned the base model using 4-bit quantization (QLoRA) with a maximum sequence length of 4096 tokens. Parameter Efficient Fine-Tuning (PEFT) was applied via Low-Rank Adaptation (LoRA) with a rank of $r=64$, a scaling factor $\alpha=32$, and a dropout rate of $0$. The LoRA adapters targeted the query, key, value, output, gate, up, and down projection modules. Training was performed for 2 epochs using the 8-bit AdamW optimizer with a learning rate of $1 \times 10^{-4}$ and a linear decay schedule with a warmup ratio of $0.03$. We utilized a per-device batch size of 2 with 8 gradient accumulation steps, resulting in an effective batch size of 16. To ensure reproducibility, the random seed was set to 3407. 

    \item \textbf{Index translation.} We use our finetuned version of the model to translate all benchmark \samples from our \database into their respective shorthand notation.   This step yields cost-efficient annotation, constrains the vocabulary, and enforces consistent skill verbalization across all the \samples in the transformed \database. Some examples from the final \database are provided below:
    { 
\small
\setlength{\tabcolsep}{2pt} 
\renewcommand{\arraystretch}{0.85} 
\begin{longtable}{
    >{\RaggedRight}p{0.10\linewidth} 
    >{\RaggedRight}p{0.68\linewidth} 
    >{\RaggedRight\arraybackslash}p{0.20\linewidth}
    }
    \toprule
    \textbf{ID} & \textbf{Test Case} & \textbf{Shorthand} \\
    \midrule
    \endfirsthead

    \toprule
    \textbf{ID} & \textbf{Test Case} & \textbf{Shorthand} \\
    \midrule
    \endhead

    \bottomrule
    \endfoot

    mathqa & 
    if $x$ is a number such that $x^2 + 2x - 24 = 0$ and $x^2 - 5x + 4 = 0$, then $x = ?$ Options: a) 4, b) -4, c) -3, d) -6, e) 1 & 
    equation\_solving \& quadratic\_equations \& solution\_id \\
    \midrule 

    triviaqa & 
    Who invaded Europe from Mongolia and Turkey over 300 years, beginning in the 13th century? & 
    factual\_recall \& historical\_figures \& mongols \\
    \midrule

    alpacaeval & 
    Provide a formula for computing the nth term in the given sequence: 5, 14, 23, 32, 41, ... & 
    equation\_solving \& sequence\_formula \& nth\_term \\
    \midrule

    arc\_c & 
    Scientists often change conclusions when new information becomes available. Which statement was changed? Options: Moon is in outer space / Plants make own food / Sun revolves around Earth & 
    reading\_comp \& analysis \& scientific\_conclusions \\
    \midrule    
\end{longtable}
} 
\end{enumerate}

Once this translated \database is generated, we create the FAISS \cite{douze2024faiss} index for this \database to use for retrieval when \shorthand is used as an anchor generation strategy. 

\subsection{Anchor Generation Prompts}
\label{promptsforbaselines}
\paragraph{\segment Baseline.}
\begin{takeaway}
Segment Generation Prompt \\

"""Given a user query, create **three brief refinements** of that query that are appropriate for retrieving examples representing the interest highlighted in the query. Use the following rules: 

- If the query is compositional, i.e., it contains multiple concepts, make different operationalizations of each concept. For example: if the query is `scientific visualizations for scientific papers` - each refinement should focus on a different sub-part of this query: like scientific visualizations, visualizations for scientific papers and tools for visualizations. 
- If the query is about some broad capability, enlist different sub-topics that test for that capability. For example: if the query is `trivia` - each refinement can include common topics included in trivia questions like `celebrity history`, `geography`, `world history`, etc. 
- If the query is about some broad topic, enlist different related domains and sub-topics that are related to that topic. For example: if the query is `biology` - each refinement can include different related domains like `genetics`, `ecology`, `dermatology`, etc.
- If the query is about a specific task, generate different contextual forms of that task. For example: if the query is `Creative Writing` - each refinement can include different types of creative writing like `poetry`, `short stories`, `novels`, etc.
- If the query describes a scenario - `Cooking indian food` - list different activities that can be performed in that scenario. For example: `indian spices`, `indian food recipes` and `indian desserts`.
- If the query is a proper Noun - `Harry Potter` - generate at least one query that highlights the broad task associated with that proper noun, for example: `fantasy literature` or `magic`, and one that lists the most popular form of that proper noun, for example, `Harry Potter Books`.
- If the query is a question - `What is the Capital of France ?` - generate at least one query that highlights the topic of that question. For example: `Capitals of different countries`, `Facts about France` and `Paris`.
- If it falls under neither of these categories, for example: `Can Elephants fly ?` - generate queries that cover reasonable aspects of that query: For example: `Animals that can fly`, `Elephant facts` and `Flying facts`.

* Make sure that one refinement is not a subset of another refinement.
* Make sure that at least one refinement highlights the broad domain of the query. For example, if the query is `chess` - one refinement should include `board games`.
* Avoid making overly complicated refinements by using simple words and phrases.
* Focus on what would attract different types of relevant test cases or documents from existing test benchmarks.
ALWAYS Return your responses in the following XML format; each refinement should be in a separate $<$refinement$>$ tag: 
$<$refinements$>$
    <refinement> </refinement>
</refinements>
DO NOT INCLUDE ANY OTHER TEXT IN YOUR RESPONSE or any reasoning in your response.
The user query is: {query}
\end{takeaway}

\paragraph{\testcase Baseline.}
\begin{takeaway} 

Testcase Generation Prompt \\
"""Given a user query, generate 3 testcases that is representative of the user's interest that can help evaluate an AI Model on its competence on the users intent. The rules for generating the testcases are as follows: 
- The testcase should be STRICTLY between 1-2 lines and not very verbose.
- Each testcase can be of a different format (Long Form, MCQ, Fill in the blank, etc.)
- Only include the question/input for a testcase and not the expected output. 
- Return each testcase wrapped in a separate $<$testcase$>$ $<$/testcase$>$ tag. Your final outputs should be of the form:
$<$testcases$>$
    $<$testcase$>$ Some question related to the query $<$/testcase$>$
    $<$testcase$>$ Some question related to the query $<$/testcase$>$
    $<$testcase$>$ Some question related to the query $<$/testcase$>$
$<$/testcases$>$
DO NOT INCLUDE ANY OTHER TEXT IN YOUR RESPONSE or any reasoning in your response.
- The user query is: \{query\}
"""

\end{takeaway}

\section{Embedding-Based Semantic Retrieval Details}
\label{app:retrievers}

\subsection{In Context Examples for the Embedding Models (used during \vanilla, \testcase and \segment Index Creation)}
\label{app:icl-embedding-prompt}

\begin{longtable}{>{\RaggedRight}p{0.25\textwidth} >{\RaggedRight}p{0.65\textwidth}}
\toprule
\textbf{\usecase} & \textbf{Relevant \sample} \\
\midrule
\endfirsthead

\multicolumn{2}{c}{\tablename\ \thetable\ -- \textit{Continued from previous page}} \\
\toprule
\textbf{Query Activity} & \textbf{Response / Relevant Record} \\
\midrule
\endhead

\midrule
\multicolumn{2}{r}{\textit{Continued on next page}} \\
\endfoot

\bottomrule
\endlastfoot

Verify scientific claims & Claim: AdaBERT achieves inferior performance while significantly worsening the effiency by 12.7x. Evidence: The dataset contains 3772 word pairs. The accuracy of ConVecs 70.02\% is not significantly different from the accuracy of SimDiffs (72.4\%). Label: Negative/Refuted. \\

Blogging & Write a blog announcing the opening of a new mall in the town of Serenity. \\

Unit Testing Code & Question: An \textbf{anagram} is the result of rearranging the letters of a word to produce a new word. \textbf{Note:} anagrams are case insensitive Complete the function to return \texttt{true} if the two arguments given are anagrams of each other; return \texttt{false} otherwise. \textbf{Examples:} foefet is an anagram of toffee. ut\_id: 0 code - import unittest class TestAreAnagramsFunction(unittest.TestCase): This class contains unit tests for the are\_anagrams function. def test\_anagram(self): Test that two anagrams return True. self.assertTrue(are\_anagrams("foefet", "toffee")) def test\_not\_anagram(self): Test that two non-anagrams return False. \\

Star Gazing & The Big Dipper is a \_\_\_\_\_\_ constellation. \\

Resume for Machine Learning Engineer & I am a software engineer with 5 years of experience in the industry. Can you help me write a resume? \\

\end{longtable}

\vspace{1em}
\noindent\textit{Note: All records share the instruction: "Given an activity, retrieve relevant records that have the same underlying concepts and topics as the given activity"}

\subsection{Comparing Embedding-Models and Retriever Configurations for the Retriever Backbone}
\label{retriever-comparison}

We measure the \systemp and \recall on the raw user \usecases from the \topic on several widely used dense retriever backbones spanning (i) standard BGE bi-encoders: \href{https://huggingface.co/BAAI/bge-base-en-v1.5}{BAAI/bge-base-en-v1.5} and \href{https://huggingface.co/BAAI/bge-large-en-v1.5}{BAAI/bge-large-en-v1.5} which are popular semantic embedding construction models; (ii) \href{https://huggingface.co/BAAI/llm-embedder}{BAAI/llm-embedder}, a popular instruction-tuned/LLM-derived embedding model suited for diverse retrieval augmented generation applications especially involving long, abstract queries \cite{zhang-etal-2024-multi-task}; and (iii) a cross-encoder reranker \href{https://huggingface.co/BAAI/bge-reranker-base}{BAAI/bge-reranker-base}, which jointly models query–\sample interactions rather than independent embeddings. The reranker's inclusion also allows us to test whether higher compute per query can compensate for the representation gap induced by broad, under-specified practitioner \usecases. Finally, we include the (iv) \href{https://huggingface.co/BAAI/bge-en-icl}{BAAI/bge-en-icl}, an embedding construction model that is designed to better encode instructional intent and task structure using guidance from In-Context Learning examples demonstrating ideal recall. We leverage the latter to diversify our retriever's recall of diversely formatted \samples by including the same in the in-context learning examples we generate for the case. \S\ref{app:icl-embedding-prompt} outlines the in-context examples used when generating the embeddings for the \texttt{BAAI/bge-en-icl} model. All values reported here at $k=20$.

\begin{table}[h]
\centering
\small
\begin{tabular}{lcccc}
\hline
Retriever Type & \systemp & \recall & NDCG@10 / MRR & Processing Time (s) \\
\hline

\texttt{bge-base-V1.5} & 0.208 & 0.004 & 0.291 & 0.048 \\
\texttt{bge-large-V1.5} & 0.332 & 0.004 & 0.353 & 0.044 \\
\texttt{llm-embedder} & 0.488 & 0.012 & 0.537 & 0.029 \\
\texttt{bge-en-icl} & \textbf{0.694} & \textbf{0.218} & \textbf{0.684} & 1.040 \\
\texttt{bge-reranker} & 0.664 & 0.176 & 0.632 & 14.030 \\
\hline
\end{tabular}
\caption{Retriever Performance Comparison: \texttt{bge-en-icl} achieves the highest \systemp, \recall and \ndcg with reasonable latency.}
\label{tab:retriever_comparison}
\end{table}

Table~\ref{tab:retriever_comparison} shows that \texttt{bge-en-icl} provides the best overall trade-off: it achieves the highest system precision (\systemp=0.694), \recall (0.218), and ranking quality \ndcg, substantially outperforming both generic BGE encoders and the stronger \texttt{llm-embedder} baseline (\systemp=0.488, \recall=0.012). Notably, while the \texttt{bge-reranker} is competitive in accuracy (\systemp=0.664, recall=0.176), its latency is an order of magnitude higher (14.03s vs.\ 1.04s), making it impractical as a default backbone for the interactive needs for our tool. In contrast, \texttt{en-icl} yields the strongest effectiveness metrics while remaining within a reasonable per-query compute budget, and we therefore adopt \texttt{bge-en-icl} as our default retriever backbone in \methodname.

\clearpage
\section{Filtering selection prompt}
\label{app:selection_prompt}

\begin{takeaway}
SELECTION PROMPT: You are evaluating whether a test case is relevant for assessing an AI model's ability to help with a specific user task. You are given:
- $<$user\_intent$>$ Some topic, a description of some scenario, a capability or an application that some user is interested in $<$user\_intent$>$
- $<$test\_case$>$ Some question/test case that might be used to evaluate that system. $<$/test\_case$>$.

A test case is RELEVANT if correctly answering it would demonstrate capabilities that directly support the user's intent. Consider the following:
\begin{itemize}
    \item \textbf{Knowledge Overlap}: Does the test require knowledge domains that overlap with what the user needs?
    \item \textbf{Skill Transfer}: Would the reasoning, analysis, or problem-solving skills needed for this test directly apply to the user's task? 
    \item \textbf{Practical Utility}  If an AI can handle this test case well, would that increase confidence it can help the user with their specific goal?
\end{itemize}
Rate the relevance as one of the following:
\begin{itemize}
    \item 1: Test case is relevant to the user's intent.
    \item 0: Test case maybe relevant to the user's intent but not directly relevant.
    \item -1: Test case is not relevant to the user's intent
\end{itemize}
Provide your rating (only 1,0 or -1) wrapped in the $<$score$>$ and $<$/score$>$ tags.
\end{takeaway}

\section{Evaluation Setup and Datasets}
\label{app:evaluation-setup}
\subsection{Benchmarks Included in \methodname}
\label{app:benchmarks}

\begin{table}[htbp]
\small
\centering
\caption{Benchmarks Used for Example Retrieval}
\label{tab:benchmarks}
\begin{tabular}{ll|ll}
\toprule
\textbf{Benchmark} & \textbf{Citation} & \textbf{Benchmark} & \textbf{Citation} \\
\midrule
\texttt{IFEval} & \citet{zhou2023instruction} & \texttt{logiqa} & \citet{liu2021logiqa} \\
\texttt{MMLU} & \citet{hendrycks2020measuring} & \texttt{mathqa} & \citet{amini2019mathqa} \\
\texttt{TruthfulQA} & \citet{lin2022truthfulqa} & \texttt{openbookqa} & \citet{mihaylov2018can} \\
\texttt{HellaSwag} & \citet{zellers2019hellaswag} & \texttt{qasper} & \citet{dasigi2021dataset} \\
\texttt{ARC} & \citet{chollet2019measure} & \texttt{triviaqa} & \citet{joshi2017triviaqa} \\
\texttt{Winogrande} & \citet{sakaguchi2020winogrande} & \texttt{webqs} & \citet{berant2013semantic} \\
\texttt{BBH} & \citet{suzgun2023challenging} & \texttt{AlpacaEval} & \citet{dubois2024length} \\
\texttt{gsm8k} & \citet{cobbe2021training} & \texttt{ComplexBench} & \citet{wen2024benchmarking} \\
\texttt{headqa} & \citet{vilares-gomez-rodriguez-2019-head} & \texttt{Infobench} & \citet{qin2024infobench} \\
\texttt{humaneval} & \citet{chen2021evaluating} & & \\
\bottomrule
\end{tabular}
\end{table}
\subsection{Evaluation Categories}
\label{evaluation-categories}
\label{app:novel-testset-construction}

\paragraph{Motivation} Skill-based testing is directly aligned with the traditional motivation for constructing LLM evaluation benchmarks, so practitioners directly benefit from being able to use \methodname's analysis to anticipate the need to construct a new skill-oriented benchmark. 
Topics and application-oriented testing is more implicitly relevant to discovering new facets where evaluating LLMs could help discover new strengths and weaknesses. These set of queries can help practitioners decide new contexts to include in future benchmarks which are increasingly focusing on maximizing coverage and topical diversity of real-world user intents.

\paragraph{\textit{Topics} use-cases.}
An index is constructed using only BIG-bench Hard (BBH) and MMLU test sets, covering 58 use-cases (tasks and subjects). 
MMLU college-level subsets are excluded due to topic redundancy (e.g., ``Human Aging'', ``High School Biology'', ``Clinical Knowledge''). 
Each topic-based query tests for focused conceptual alignment on a single subject area, enabling controlled measurement of retriever precision on narrow topical domains.

\paragraph{\textit{Skills} use-cases.}
To represent practitioner-facing evaluation intents, we derive 50 skill-based use-cases from the Clio analysis of real-world model usage~\citep{tamkin2024clio}. 
High-level skills (e.g., ``instruction-following'', ``critical thinking'', ``creative writing'') are prompted to Claude-3-7-Sonnet for naturalistic phrasing. 
The evaluation index aggregates $\sim$67k examples from 20 modern benchmarks such as IFEval~\citep{zhou2023instruction}, InfoBench~\citep{qin2024infobench}, MMLU, and BBH, enabling retrieval testing over both long-form and MCQ styles. 
This setup evaluates whether the retriever can generalize across benchmark formats and domains.

\paragraph{\textit{Applications} use-cases.}
These 50 Clio-seeded synthetic prompts simulate concrete tasks like ``studying for the SAT'' or ``debugging HP printer errors.'' 
They assess how well retrievers locate evaluation items relevant to practical scenarios. 
Precision@20, NDCG@20, and yield are reported per application, using the same 20-benchmark index as the skills setup. 
Such testing is crucial for understanding real-world generalization and identifying missing evaluation coverage across practical tasks.

\paragraph{\textit{Novel} use-cases.}
We circulated a Google Form surveying NLP practitioners (graduate students within AI-focused field) to seek novel \usecases that they were interested in evaluating LLMs on. The participants were allowed to make more than one submission and up to 3 submissions. Each participant was requested to provide at least one peer-reviewed research paper that demonstrates the common utility of the \usecase of their interest to the broader scientific community, ensuring the relevance and validity of our evaluation scenarios. This generated a set of 21 \usecases. Some \usecases were processed to omit personal identifiers - for example,`recommend me food that can satisfy my dietary constraints` was modified to `recommending food that satisfies dietery restrictions`.

\subsection{Prompts for Generating Usecases}
\label{app:use-case-generation}

We use the Clio \cite{tamkin2024clio} paper to seed the usecases for which we test \methodname. This is done to ensure that we don't pick usecases that are too disconnected from real-world needs.

\paragraph{Generating Usecases for the \skill}
For generating the skill set we use the high-level task categories emerging from a random sample of 1M Claude.ai conversations as described in the paper (Table 7, Appendix F in \cite{tamkin2024clio}).   
 
\begin{takeaway}
    You are given a list of high-level use case categories for which Claude is used by real-world users. Your task is to identify the underlying fundamental skills that users need when engaging in tasks in these categories.

    For each category, think about:
    \begin{itemize}
        \item What are the core cognitive abilities required? (e.g., reasoning, analysis, creativity)
        \item What technical competencies are involved? (e.g., coding, data manipulation, system design)
        \item What interpersonal or communication skills are needed? (e.g., persuasion, empathy, cultural awareness)
        \item What metacognitive or strategic capabilities are at play? (e.g., planning, decision-making, problem diagnosis)
    \end{itemize}

    Generate a list of 50 broad, transferable skills that represent these fundamental capabilities. These skills should be:
    \begin{itemize}
        \item \textbf{Atomic}: Each skill should represent a single, measurable capability
        \item \textbf{Cross-domain}: Skills should be applicable across multiple use case categories
        \item \textbf{Evaluable}: Each skill should be something that can be tested or benchmarked
        \item \textbf{Granular enough to be distinctive}: Avoid overly generic terms like "intelligence" or "ability"
    \end{itemize}

    Return your answer wrapped in XML tags.
    
    Example: \texttt{<skill>cultural adaptation</skill>}
    
    \textbf{Input - Use Case Categories:}
    \begin{itemize}
        \item Web and Mobile App Development
        \item Content Creation and Communication
        \item Academic Research and Writing
        \item Education and Career Development
        \item Advanced AI/ML Applications
        \item Business Strategy and Operations
        \item Language Translation
        \item DevOps and Cloud Infrastructure
        \item Digital Marketing and SEO
        \item Data Analysis and Visualization
    \end{itemize}
\end{takeaway}

\paragraph{Generating Usecases for the \application}
For the \application, we use the ground-truth categories from an evaluation dataset of 19,476 synthetic chat transcripts that outline more specific contexts under which Claude is utilized. We ask \texttt{Claude-3.4-Sonnet} to generate applications on the basis of the provided categories using the following prompt.

\begin{takeaway}
    You are given a list of broad categories of tasks for which Claude is used by real-world users. Create a list of 50 applications based on the given categories. Do not make the compositional applications i.e., applications that combine too many tasks. Return your answer wrapped in XML tags "usecase" "usecase"
    Example: "usecase" improving home energy efficiency "usecase"
    Example: "usecase" recommending Netflix series for me "usecase"
\begin{itemize}
    \item Software development questions
    \item Elementary school homework help
    \item Technology troubleshooting
    \item Health and fitness advice
    \item Questions about geopolitics
    \item Parenting and childcare tips
    \item Language learning and translation help
    \item Financial planning and investment
    \item Theological and philosophical questions
    \item Environmental science and sustainability
    \item Book discussions and literary analysis
    \item Sports rules and strategy questions
    \item Cooking and recipe inquiries
    \item Job application questions
    \item Home improvement and DIY projects
    \item Pet care and animal behavior
    \item Romantic relationship advice
    \item Movie and TV show recommendations
    \item Music theory and instrument learning
    \item Tourism and travel questions
\end{itemize}
\end{takeaway}

\subsection{\known \usecases and their corresponding Gold Test Sets}
Table \ref{tab:usecase_mapping} specifies the \usecases designed for the \known. 

\label{promptwithgoldtestsets}
\begin{table}[htbp]
\centering
\small
\begin{tabular}{p{11cm}p{4cm}}
\hline
\textbf{Usecase} & \textbf{Gold Test Set} \\
\hline
Learn how to measure my carbohydrate intake and make my meals more balanced & nutribench \cite{Hua2024NutriBenchAD} \\
\hline
Test my knowledge of different sports and their respective players & bb\_sports\\
\hline
Help me learn more about the Hindu religion & bb\_hindu  \\
\hline
Test me on different topics in science & sciq \cite{welbl-etal-2017-crowdsourcing}\\
\hline
Help me answer questions for my biomedical research & pubmed \cite{jin2019pubmedqa} \\
\hline
Help me mitigate different biological, cyber, and chemical attacks & wmdp \cite{li2024wmdp} \\
\hline
Explain the ethically right choice to make under different kind of situations & moral\_stories \cite{emelin-etal-2021-moral} \\
\hline
Can help me write one-line python codes & bb\_code\_line \\
\hline
Helps me find common substrings between strings & bb\_cs\_algo \\
\hline
Assists me in fact checking my claims & bb\_fact\_checker \\
\hline
Tests me on the properties of human organs & bb\_human\_organs  \\
\hline
Teaches me how to use metaphors in daily language & bb\_identify\_odd \\
\hline
Learn how to identify common misconceptions & bb\_misconceptions  \\
\hline
Helps me resolve moral dilemmas & bb\_moral\_permissibility \\
\hline
Teaches me how to identify the underlying concept or theme among a few things & bb\_novel\_concepts  \\
\hline
Can solve riddles for me & bb\_riddle\_sense \\
\hline
Can help me understand the social implications of my actions & bb\_social\_iqa \\
\hline
Help me figure out if someone supports my argument & bb\_social\_support \\
\hline
Tests my emotional intelligence & bb\_strange\_stories \\
\hline
\end{tabular}
\caption{Prompts with Corresponding Gold Datasets used for the  \textit{Known Validation} Set. All datasets prefixed with ``bb'' are included in the Big Bench Dataset \cite{srivastava2023beyond}.}
\label{tab:usecase_mapping}
\end{table}

\subsection{Evaluation Metrics}
\label{app:eval-metrics}

We use the following metrics for evaluation: 

\begin{enumerate}[label=(\alph*)]
    % \item \textbf{Yield}: Measures how many relevant samples were returned for the user's use case. Formally:
    % \begin{equation}
    %     \text{Yield} = |\{d \in \mathcal{R} : \text{relevant}(d, q)\}|
    % \end{equation}
    % where $\mathcal{R}$ is the set of retrieved \samples and $q$ is the user's \usecase. This metric helps readers understand ``Are benchmarks representative of their intent?''

    \item  $\text{MethodPrecision@k}_t = \frac{|\{d_i : i \leq k \wedge S_t(d_i) \in \{1, 0\}\}|}{k}$ - This quantifies the precision of the ranking methods i.e., the number of relevant or partially relevant retrieved \samples before the relatedness filtering within \methodname. 

    \item $\text{SystemPrecision@k}_t = \frac{|\{d_i : i \leq k \wedge L_t(d_i) \in \{\text{REL}, \text{PART}\}\}|}{k}$ - This quantifies the precision of \methodname i.e., the number of relevant or partially retrieved \samples by \methodname after the relatedness filtering. 

    \item $\text{Intersection@k}_t = \frac{\text{hits}_t(k)}{k}$ For the tie-breaking experiment, we also report the $\text{Intersection@k}_t$ which helps identify which method recalls the highest number of \samples among all the \samples returned by all the methods.

    \item \textbf{Recall@k}: In the Bounded Recall setup, measures how many of the returned \samples actually belong to the expected gold test set for that use case:
    \begin{equation}
        \text{Recall@k} = \frac{|\{d \in \mathcal{R}_k \cap \mathcal{G}\}|}{|\mathcal{G}|}
    \end{equation}
    where $\mathcal{G}$ is the gold test set for the given use case.

    \item  \textbf{Common cutoff:} $K_c = \max\{k : |T_k| \geq 0.9 \cdot |T|\}$
    \item \textbf{Correlation with Model Ranking}: Wherever gold sets are available  compute the correlation between the rankings of models when evaluated on a validation set (gold test set) versus our retrieved test set of the same advertised intent. 

    \item \textbf{NDCG@k (Normalized Discounted Cumulative Gain)}: Measures the quality of ranking by considering both relevance and position, with higher-ranked relevant documents contributing more to the score:
    \begin{equation}
        \text{NDCG@k} = \frac{\text{DCG@k}}{\text{IDCG@k}}
    \end{equation}
    where DCG@k (Discounted Cumulative Gain) is defined as:
    \begin{equation}
        \text{DCG@k} = \sum_{i=1}^{k} \frac{2^{\text{rel}(d_i)} - 1}{\log_2(i + 1)}
    \end{equation}
    and IDCG@k (Ideal Discounted Cumulative Gain) is the DCG@k score of the perfect ranking. Here, $\text{rel}(d_i)$ is the relevance score of document $d_i$ at position $i$, and the logarithmic discount factor $\log_2(i + 1)$ reduces the contribution of lower-ranked documents. 
\end{enumerate}

\subsection{Evaluation Prompts}
\label{app:evaluation_prompt}

The key distinction between our \systemp and \methodp is guided by the distinction between our evaluation prompt used by the LLM Judge for the Filtering Stage (referred to as the Selection Prompt) and the LLM Judge used for the final evaluation. While the \methodp evaluation aims at ascertaining which proportion of \samples ranked by any anchoring method (\testcase, \shorthand, etc) are related to the user's intent, the \systemp evaluation prompt is strongly situated in a practitioner's intent of actually using the retrieved \samples for a downstream task. We make this distinction assuming that practitioners might be lenient in claiming relatedness but not necessarily find a testcase useful for their specific intent if the test case is only broadly related to their task. 

We corroborate this assumption with annotators during our human evaluation. For most comparisons, we stick to \systemp, as that is the practically useful metric that practitioners would expect maximized.

\label{app:relevance_prompt}
\begin{takeaway}
EVALUATION PROMPT: """You are evaluating whether a test case is relevant for assessing an AI model's ability to help with a specific user task. You are given:
- $<$user\_intent$>$ Some topic, a description of some scenario, a capability or an application that some user is interested in $<$/user\_intent$>$
- $<$test\_case$>$ Some question/test case that might be used to evaluate that system. $<$/test\_case$>$
Your evaluation criteria is:
A test case is RELEVANT if successfully answering it would demonstrate capabilities that directly support the user's intent. Consider the following:
\begin{itemize}
    \item \textbf{Knowledge Overlap}: Does the test require knowledge domains that overlap with what the user needs?
    \item \textbf{Skill Transfer}: Would the reasoning, analysis, or problem-solving skills needed for this test directly apply to the user's task? 
    \item \textbf{Practical Utility}  If an AI can handle this test case well, would that increase confidence it can help the user with their specific goal?
\end{itemize}
Rate the relevance as one of the following:
\begin{itemize}
    \item RELEVANT: If an AI can answer this test case correctly, it would strongly imply that it can help the user with their specific goal.
    \item PARTIAL RELEVANT: If an AI can answer this test case correctly, it would somewhat imply that it can help the user with their broad goal i.e., with the topic or the skill needed by the user. 
    \item IRRELEVANT: Answering this question, would not provide any useful indication about if the AI can fulfill the user's intent. 
\end{itemize}
Provide your rating and wrapped in the $<$label$>$ and $<$/label$>$ tags. 
\end{takeaway}

\section{Additional Evaluation of Anchor Generation Methods for Retrieval}
\label{app:additional-retrieval-eval}

\subsection{Tie-breaking: Computing Correlation between Model Performances on Retrieved Set and Gold Test Set for \topic}
\label{app:tie-breaking-correlation}
Measuring the correlation between the performance of a model on a gold test set that is explicitly designed for a \usecase with the performance of the model on the set of retrieved \samples can be a strong sanity check on the utility of the retrieved \samples. Accordingly, for the \topic, we compute the performance of $N=8$ models on the 58 BBH and MMLU testsets and then compute their correlation with the performance of the models on the sets of retrieved \samples for the same \usecases. We use the following models for our evaluation: 
We use the following eight models for our evaluation: \texttt{Llama-3.1-8B-Instruct}, \texttt{Llama-3.1-Tulu-3-8B}, \texttt{Mistral-7B-Instruct-v0.2}, \texttt{Qwen1.5-14B-Chat}, \texttt{Qwen2.5-7B-Instruct}, \texttt{c4ai-command-r7b-12-2024}, \texttt{gemma-2-9b-it}, and \texttt{gemma-3-4b-it}. Since all the benchmarks in this set use either accuracy or exact string match as their metric for reporting accuracy, we aggregate the per-sample accuracy or exact-string match ($\in{0,1}$) and then normalize by the number of retrieved \samples to obtain each model's performance on the retrieved set. As a baseline, we also plot the correlation between the performance of the models on the bottom-k and a random slice of the benchmark index as a way of calibrating random correlation. 

\begin{figure}[t]
    \centering
    \includegraphics[width=0.9\linewidth]{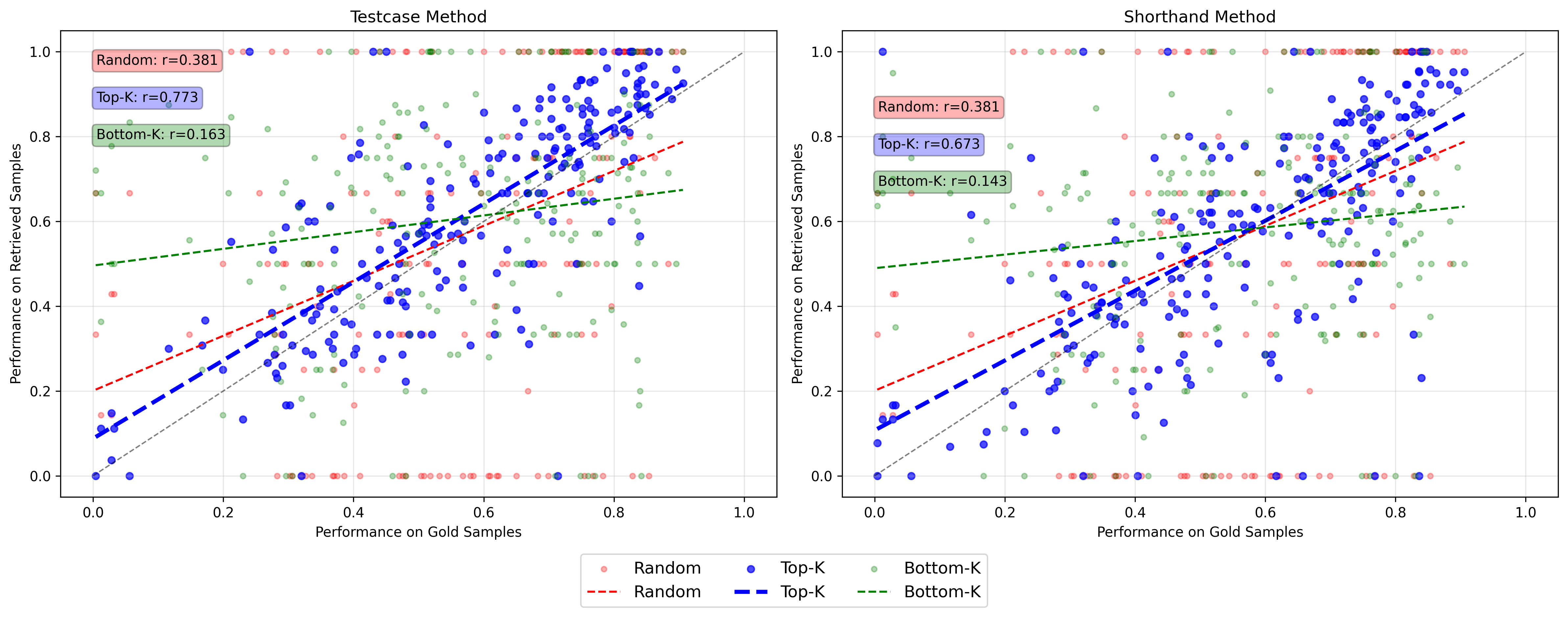}
    \caption{We observe that performance on the retrieved testset and the gold test set (BBH and MMLU for the \topic set) show reasonable to high correlation.}
    \label{fig:correlation-btw-gold-retrieved-set}
\end{figure}

Figure \ref{fig:correlation-btw-gold-retrieved-set} demonstrates that the performance of the models on the retrieved sets and gold sets have high correlations: highest for the \testcase anchor method. 

\subsection{Tie-breaking: Estimating Overlap with Approximated Gold Set for \application and \skill}
\label{app:retrieved-set-overlap}

Use-cases in the \emph{application} and \emph{skill} sets do not have pre-defined gold sets of human-curated examples.
Therefore, we cannot calculate the $\text{recall}@k$ metric for these use-cases.
Instead, we compare the different retrieval methods using a metric we call $\textit{intersection}@k$.
This metric measures the overlap between a method's relevant examples and the full set of examples that are deemed relevant by any method.
To do this, we first run each retrieval method for a relatively large value of $k=80$ and then take the union of all the \samples that are judged as relevant by any method.
Then, for a method \(m\) with ranked list \(R_m(q)=[d_1,d_2,\dots]\), we compute the intersection of the \samples up to rank $k$ retrieved by \methodname with the union set and divide by $k$.

\begin{figure}[t]
    \centering
    \small
    \includegraphics[width=0.8\linewidth]{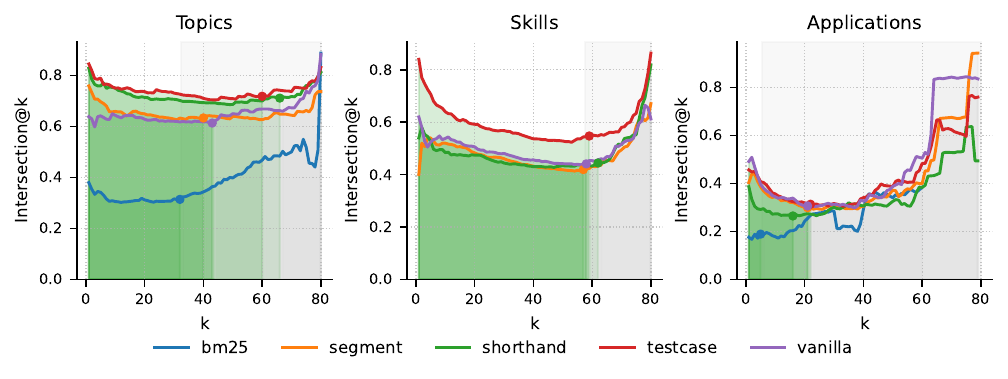}
    \caption{$\text{intersection}@k$ in approximated recall setup: Gray region indicates precision for fewer than $90\%$ of queries. For each method we compute $K_{common}$: the largest k for which $\geq 90\%$ of tasks contribute a value at every position $\leq$ k (i.e., their lists are at least that long). The area up to $K_{common}$ under that method’s curve is filled green (trusted, high task support); the tail beyond $K_{common}$ is filled grey (lower support). A dot marks ($K_{common}$, Precision at K common.)}
    
    \label{fig:approx-recall}
\end{figure}

Figure~\ref{fig:approx-recall} shows that the \textit{testcase} method consistently has higher $\text{intersection}@k$ values across all three sets of use-cases.

To investigate the cause of the spike observed at the tail end of the intersection (Figure~\ref{fig:approx-recall}), we compute the support fraction across $k$ which denotes what fraction of the queries actually contribute to the precision at a particular value of $k$.
Formally, for method $m$, the \emph{support} is $s_m(k)=|\{q:\ |G(q)|>0\ \wedge\ |R_m(q)|\ge k\}|$, and the support fraction is
$\hat{s}_m(k)=s_m(k)/Q$.
We define $K_{\text{common}}(m)=\max\{k:\ \hat{s}_m(k)\ge 0.9\}$, so points beyond
$K_{\text{common}}$ are backed by fewer than $90\%$ of queries. In Figure \ref{fig:approx-recall}: the green area hence marks the common area versus the greyed out region indicates the region where $\text{intersection}@k$ is contributed by less than 10\% of the queries; Through this we yield a more stable comparison between the macro averages at large $k$ by avoiding dominance by a small, unrepresentative subset of easy queries where achieving high precision is naively possible. Ultimately, this experiment further provides evidence of \testcase' reasonable proficiency and generality in being able to (a) retrieve relevant \samples and (b) retrieve the largest set of relevant \samples for any \usecase. 

\subsection{Precision of \methodname as K is increased}
\label{app:precision-as-k-increases}
\begin{figure}[htbp]
\small
\centering
\includegraphics[width=0.45\textwidth]
{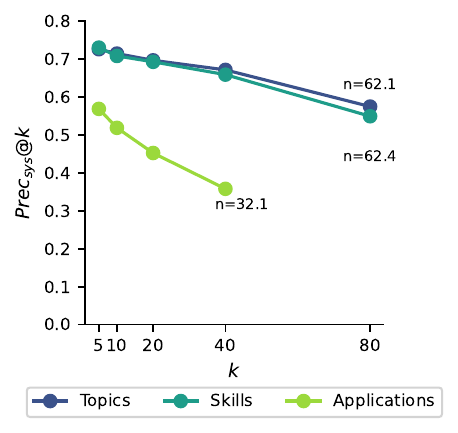}
    \caption{
    System precision (\systemp) as a function of $k$ for \topic, \skill, and \application.
    For \topic and \skill, precision remains high as $k$ increases, while \application exhibits a sharper decline.}
    \vspace{-10pt}
    \label{fig:systemp_across_k}
\end{figure}

Figure~\ref{fig:systemp_across_k} shows \systemp for the best-performing anchor (\testcase) as we vary the number of retrieved \samples, $k$.
For \topic and \skill, precision remains relatively stable as $k$ grows, indicating that \methodname\ can return larger pools of examples without rapidly flooding practitioners with irrelevant items.
This may be especially useful when practitioners want to sample a broader, larger set of \samples for using doing fine-grained selection over or using as an alternate test set for custom \usecases for which gold test sets don't exist.

In both experiments, performance on the \application set is noticeably lower across all methods (Table~\ref{tab:unified_metrics}, Figure~\ref{fig:systemp_across_k}), reflecting the difficulty of retrieving \samples for highly contextual \usecases from existing benchmarks.
This drop in precision is not surprising since \application \usecases often bundle multiple skills and entities (e.g., domain, user role, format, constraints), only some of which may be followed or \textit{available} in the \samples included in the benchmarks. 
Despite this, \testcase has high performance in this set too where the non-rewrite based \vanilla baseline actually becomes the most competitive retrieval method. In alignment with prior work, this parity between the \vanilla and \testcase based anchoring highlights the complexity in devising and choosing a single rewrite method that can maintain retrieval performance across a wide variety of practitioner \usecases \citep{kalra2025mor,salemi2024optimization}. 
\label{humanevaluation-details}

\section{Automatic Judge Calibration with Another Automatic Judge}
\label{app:automatic-judge-calibration}

To further validate the sanity of our automatic relevance judgments we also replicate the automatic precision computation in Table \ref{tab:unified_metrics} with another automatic judge i.e., \texttt{gemini-2.5-flash-lite} and report the spearman correlation between the scores enlisted by this model and our designated judge, \texttt{gpt-5-mini}.

\begin{table}[h]
\centering
\small
\begin{tabular}{lccc}
\toprule
Category & Score & \systemp [k=10] & \systemp [k=20] \\
\midrule
Overall & 0.629\textcolor{blue}{$^{\ddagger}$} & 0.700\textcolor{blue}{$^{\ddagger}$} & 0.795\textcolor{blue}{$^{\ddagger}$} \\
\midrule
Applications & 0.613\textcolor{blue}{$^{\ddagger}$} & 0.738\textcolor{blue}{$^{\ddagger}$} & 0.725\textcolor{blue}{$^{\ddagger}$} \\
Known Validation & 0.549\textcolor{blue}{$^{\ddagger}$} & 0.485\textcolor{blue}{$^{\ddagger}$} & 0.604\textcolor{blue}{$^{\ddagger}$} \\
Skills & 0.603\textcolor{blue}{$^{\ddagger}$} & 0.626\textcolor{blue}{$^{\ddagger}$} & 0.747\textcolor{blue}{$^{\ddagger}$} \\
Topics & 0.576\textcolor{blue}{$^{\ddagger}$} & 0.562\textcolor{blue}{$^{\ddagger}$} & 0.622\textcolor{blue}{$^{\ddagger}$} \\
\bottomrule
\end{tabular}
\caption{Spearman correlation coefficients. \textcolor{blue}{$^{\ddagger}$} indicates $p < 0.001$; \textcolor{red}{$^{\dagger}$} indicates $p < 0.05$.}
\label{tab:spearman_correlations}
\end{table}

\begin{figure}
    \centering
    \small
    \includegraphics[width=0.5\linewidth]{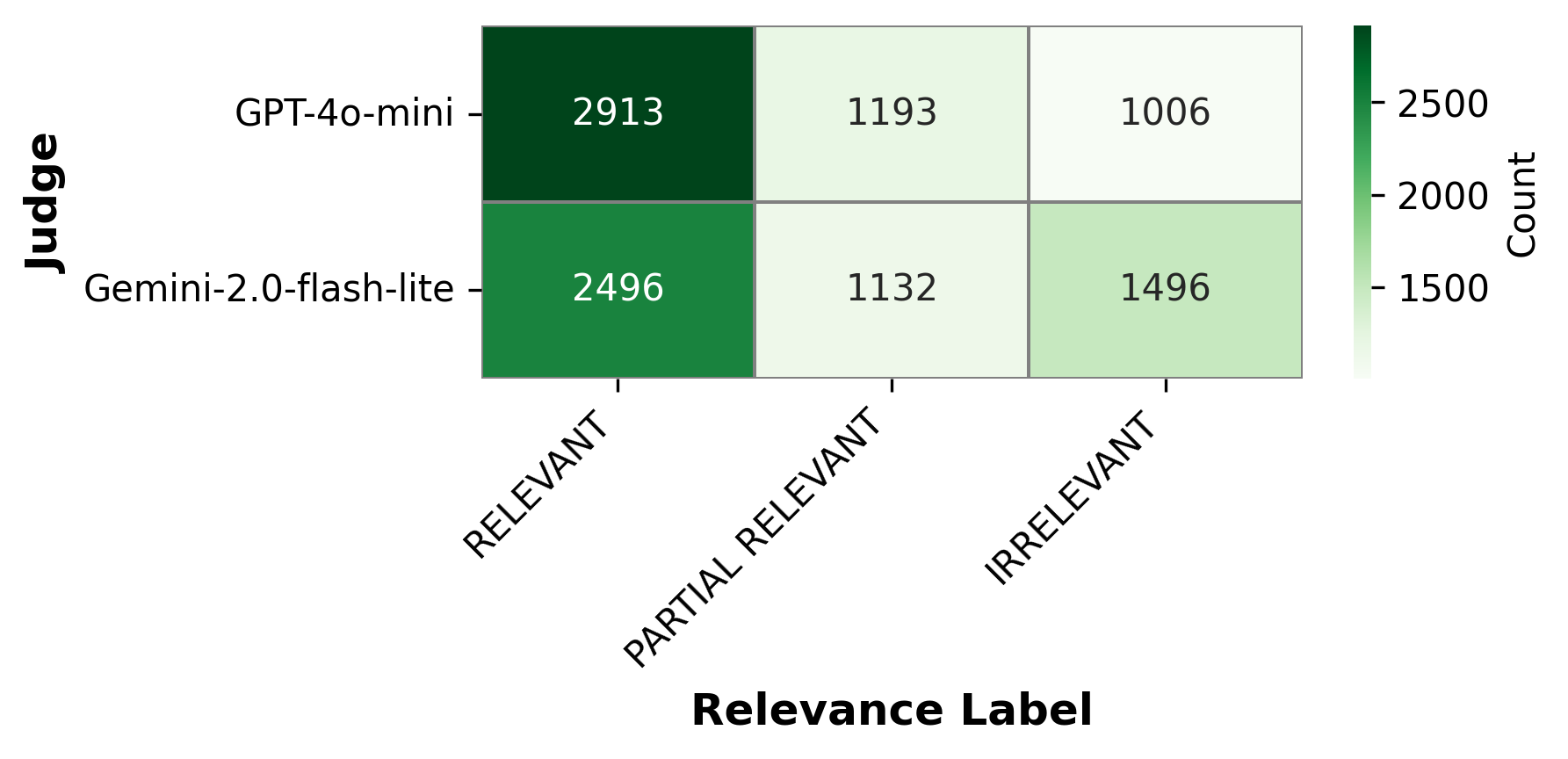}
    \caption{Marginal relevance-label distributions assigned by GPT-5-mini (designated judge) and Gemini-2.5-flash-lite over the same evaluated top-$k$ retrieved items.}
    \label{fig:judge_label_dist}
\end{figure}

The resulting Spearman correlations are consistently positive and statistically significant (Table~\ref{tab:spearman_correlations}), indicating that items judged as more relevant by our designated judge also tend to be judged as more relevant by the alternative judge. This provides evidence that the relative relevance ordering induced by our automatic evaluation is stable under judge substitution, supporting the robustness of our aggregate retrieval comparisons. 
 Figure~\ref{fig:judge_label_dist} also reports the marginal label distributions assigned by each judge across all evaluated items. Both judges use all three labels and share the same qualitative pattern---\textsc{Relevant} is most frequent, followed by \textsc{Partial Relevant} but they differ in strictness: \texttt{gpt-5-mini} assigns \textsc{Relevant} more often ($\approx$57\%) and \textit{gemini-2.5-flash-lite} assigns fewer \textsc{Relevant} labels ($\approx$49\%).  Together with the strong item-level rank agreement, this suggests that the judges largely agree on which items are comparatively more vs.\ less relevant, while disagreeing primarily in the decision boundary between \textsc{Relevant} and \textsc{Irrelevant} (i.e., Gemini is more conservative). This pattern is consistent with \emph{reasonable calibration} for our use case as aggregate precision trends seem stable (at least across method) under an alternate judge.

\section{System Infrastructure and Profiling}
\label{app:infrastructure}

\subsection{Hosting Infrastructure}
\label{app:hosting-infrastructure}

\methodname is deployed as a Google Cloud Run service in region \texttt{northamerica-northeast1}, provisioned with \textbf{4\,GiB RAM} and \textbf{1 vCPU}. To reduce cost, we set the minimum number of instances to \textbf{0}, so requests may incur cold-start overhead; each request has a \textbf{120\,s} deadline (requests exceeding this limit are terminated). Cloud Run request concurrency is set to 80.

To avoid GPU-backed embedding computation, we outsource embeddings to the \href{https://deepinfra.com}{DeepInfra} API, using \texttt{BAAI/bge-en-icl} at a rate of \textbf{\$0.0100/Mtoken}. Our reported end-to-end latency includes the external embedding API call (network + inference) in addition to backend processing.

\subsection{Latency Profiling}
\label{app:latency-profiling}

\begin{figure}[htbp]
    \centering
    \small
    \begin{subfigure}[t]{0.49\textwidth}
        \centering
        \includegraphics[width=0.9\linewidth]{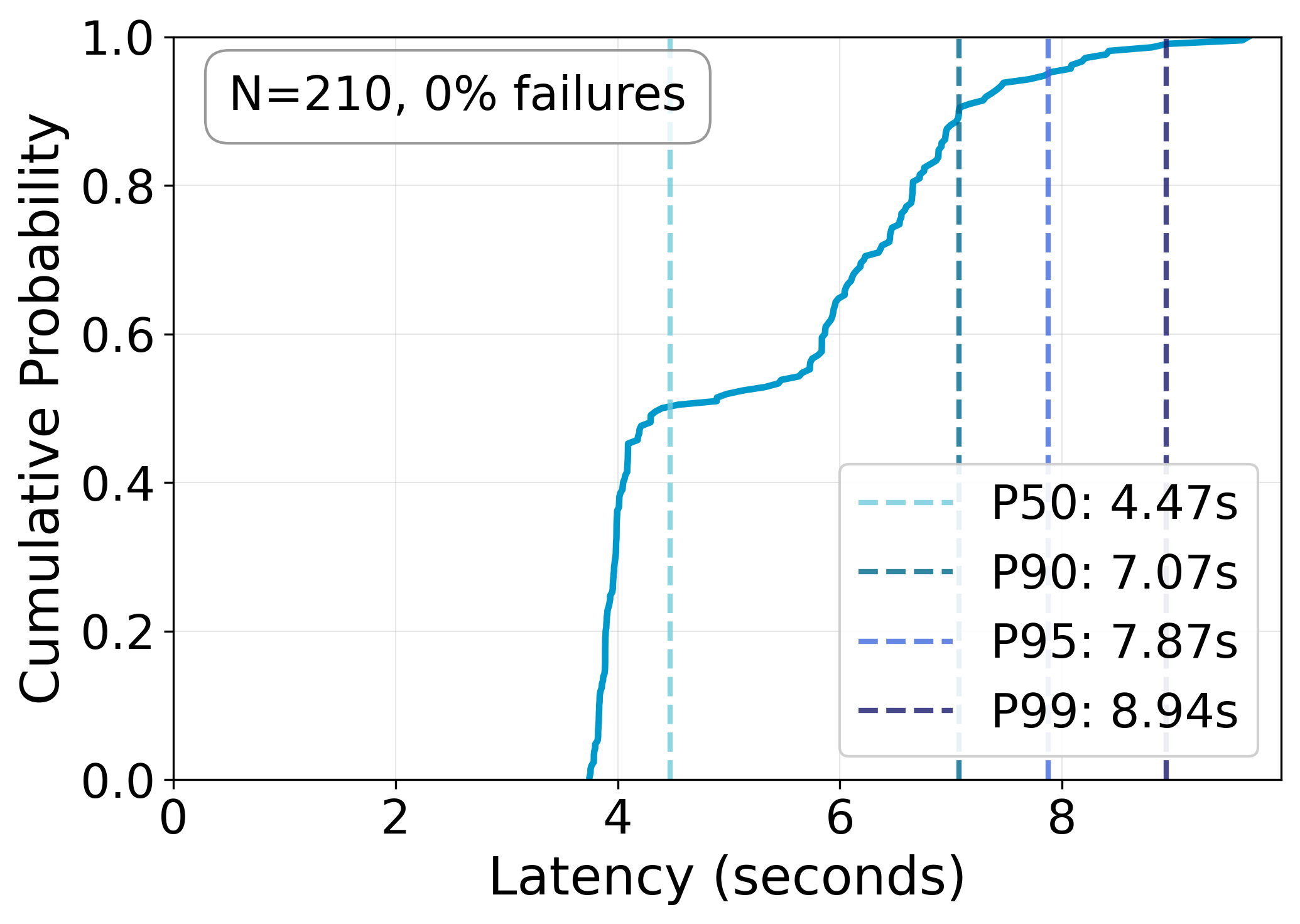}
        \caption{\textbf{End-to-end latency distribution.}}
        \label{fig:latency_cdf}
    \end{subfigure}\hfill
    \begin{subfigure}[t]{0.49\textwidth}
        \centering
        \includegraphics[width=0.9\linewidth]{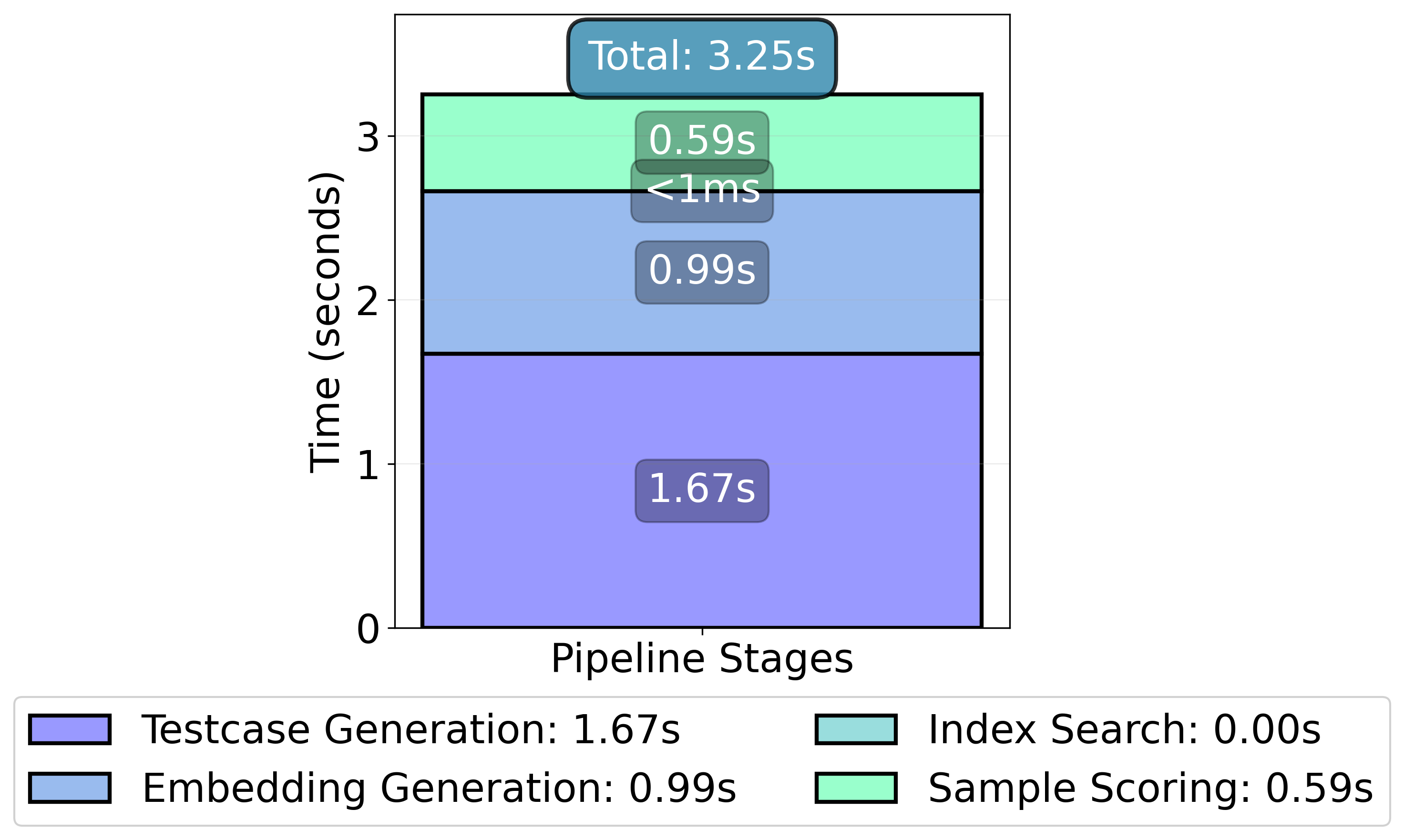}
        \caption{\textbf{Mean stage-time breakdown.}}
        \label{fig:latency_stages}
    \end{subfigure}
    \caption{\textbf{Latency of the \methodname web demo.} (Left) Empirical CDF of request latency over 210 runs of the public \methodname demo ($k{=}35$), with percentile markers (p50/p90/p95/p99). Median latency is 4.47s; tail latency remains below 8.95s at p99 (0 failures). (Right) Average time spent in each pipeline stage, showing testcase generation (1.67s; 31.7\%), embedding generation (0.99s; 18.8\%), and sample scoring (0.59s; 11.2\%) as the dominant contributors; the remaining time corresponds to end-to-end overhead.}
    \label{fig:latency}
\end{figure}

We report end-to-end user-perceived latency for the public \methodname demo over 210 requests (1 iteration per query; no failures). Latency is 5.28 ± 1.48 s (mean ± stdev), with p50 = 4.47 s, p90 = 7.08 s, p95 = 7.90 s, and p99 = 8.95 s (min 3.75 s, max 9.68 s). A stage-level breakdown attributes 1.67 s (31.7\%) to testcase generation, 0.99 s (18.8\%) to embedding generation, and 0.59 s (11.2\%) to sample scoring; the remaining ~2.02 s (~38\%) is unattributed overhead (primarily request/response handling, serialization, networking, and other pipeline glue), which we include in the end-to-end measurement.

\section{Human Evaluation of \methodname}
\label{app:humanevaluation}

We obtained an Institutional Review Board (IRB) exemption under the 2018 Common Rule (45 CFR § 46.104(d)) to conduct our study with $N=10$ annotators. Annotators were compensated with a DoorDash food order (up to \$20 USD) for approximately one hour of participation. Eligibility criteria required that participants be at least 18 years old, fluent English speakers, currently residing in the United States, and have prior experience working on at least one NLP project.

\subsection{Detailed Analysis of Human–Automatic Disagreement}
\label{app:judge-error-analysis}
\label{sec:llm-judge-validation}

\paragraph{Known Validation outliers}
A closer analysis of the \texttt{Known Validation} set reveals that the large mean difference between human and automatic relevance scores (Table~\ref{tab:judge-validation-summary}) is driven by six \usecases\ with unusually large per-\usecase\ deltas (0.40–0.75), with humans consistently assigning lower relevance.
These \usecases\ are drawn from benchmarks that test abstract or esoteric capabilities, making it difficult to “retrofit” a concise practitioner-style \usecase.
For instance, the most extreme deviation (mean difference 0.75) arises from BIG-bench’s \href{https://github.com/google/BIG-bench/tree/main/bigbench/benchmark_tasks/novel_concepts}{\texttt{novel\_concepts}} task \citep{srivastava2023beyond}, whose original goal is to
\emph{``measure the ability of models to uncover an underlying concept that unites several ostensibly disparate entities, which hopefully would not co-occur frequently.''}
We simplify this to a retrieval \usecase\ of \emph{``identifying the underlying concept for a given set of items.''}
Annotator comments (Table~\ref{tab:annotator-subjectivity}) indicate that, without the full benchmark context, this simplified \usecase\ encourages annotators to form narrow, idiosyncratic interpretations.
\samples\ that are consistent with the original benchmark designer’s intent, but not with the annotator’s inferred interpretation, are then marked as \irrelevant, increasing the human–automatic gap.
We observe similar issues for other abstract tasks in this set, suggesting that retrofitting highly specialized benchmarks into short \usecases\ can systematically bias human judgments downward.

\paragraph{Low inter-annotator agreement}
\definecolor{nude}{HTML}{228B22}
\begin{table*}[htb]
\centering
\small
\caption{Comments from Annotators over samples where we observe disagreement: Annotator A is always the annotator assigning a high relevance (1) and Annotator B is always the annotator assigning a low relevance (0).}
\label{tab:annotator-subjectivity}
\begin{tabularx}{\textwidth}{>{\raggedright\arraybackslash}p{5.5cm} >{\raggedright\arraybackslash}X}
\toprule
\multicolumn{2}{l}{\textbf{Sample} \hspace{4cm} \textbf{Annotator Comment}} \\
\midrule
\rowcolor{nude!50}
\multicolumn{2}{l}{\textbf{\usecase: Learn Differential Equations}} \\ \hline 
Prompt: Antiderivar la siguiente función f(x)=(x+1)(2x-1) & 
A says: ``solving DEs requires computing antiderivatives;'' 
B says: ``The question is testing both translation and maths;within maths, differential equation is not explicitly tested so on average I this isn't relevant''. \\
\midrule
\rowcolor{nude!50}
\multicolumn{2}{l}{\textbf{\usecase: Detecting emotional support in conversations}} \\ \hline
Group chat excerpt: Owner: @Team - Aren't you too unpunctual? We agreed on 10 o'clock yesterday ..... 

\textbf{Prompt}: Determine the emotion of the owner's last message. Choose from [negative, neutral, positive]. & 

A says : The task is sentiment classification which is relevant to detecting emotions. 

B says: The task description is specifically asking for cases that test emotional support, while the question tests emotion classification with predefined categories which aren't relevant to emotional support. \\
\midrule
\rowcolor{nude!50}
\multicolumn{2}{l}{\textbf{\usecase: Identifying the underlying concept for a given set of items}} \\ \hline
All of the following support the endosymbiotic theory .... \textit{EXCEPT}
\begin{itemize}
    \item Mitochondria and chloroplasts divide  ... 
    \item Mitochondria and chloroplasts have ribosomes ....
\end{itemize} & 
Annotator B: Given the broad nature of "concepts, factual knowledge," technically anything in a domain could share underlying concepts. This seems too abstract and insufficiently related to the specific task so I can consider this partially relevant at best".
\textit{Annotator A marks as 'relevant' but does not justify in the comments.}\\
\bottomrule
\end{tabularx}
\end{table*}

\begin{figure}[htb]
 \centering
\small
\begin{subfigure}[b]{0.48\textwidth}
    \centering
    \includegraphics[width=0.7\textwidth]{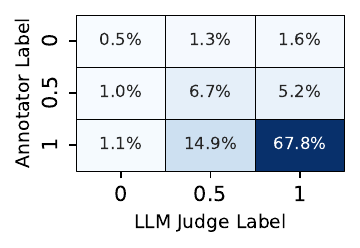}
    \caption{LLM vs Human}
    \label{fig:llm-human-correlation}
\end{subfigure}
\hfill
\begin{subfigure}[b]{0.48\textwidth}
    \centering
    \includegraphics[width=0.7\textwidth]{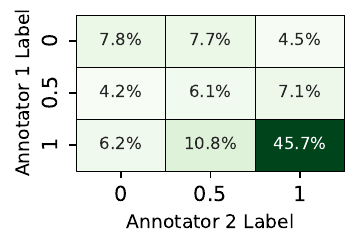}
    \caption{Human vs Human}
    \label{fig:human-human-correlation}
\end{subfigure}
\caption{Intra and Inter-Label distributions between LLM and human annotators on the \samples evaluated from the \textit{topics, skills and applications} set. Most disagreements occur between \emph{partially relevant} and \emph{relevant} categories rather than polar opposites, implying calibration rather than conceptual divergence.}
\label{fig:label-distribution-comparison}
\vspace{10pt} 
\end{figure}

We also investigate the relatively low Fleiss’s $\kappa$ for several sets (Table~\ref{tab:judge-validation-summary}).
Figure~\ref{fig:label-distribution-comparison} plots confusion matrices for (a) the LLM judge versus the majority human label and (b) pairs of human annotators, restricted to \samples\ from the \topic, \skill, and \application sets.
In both cases, most disagreements occur between \relevant\ and \partially\ relevant labels rather than between \relevant\ and \irrelevant, indicating that annotators and the LLM generally agree on whether an item is “in the ballpark” and differ mainly on how strict to be.
Table~\ref{tab:annotator-subjectivity} illustrates this with example comments from annotators on disputed items: both annotators often articulate plausible but different interpretations of what counts as “relevant” to a given \usecase\ (e.g., whether sentiment classification suffices for “detecting emotional support in conversations”).
This pattern matches prior findings that inter-annotator agreement is modest for inherently subjective constructs such as quality, relevance, or interest \citep{adams2023sparse,DBLP:conf/hcomp/McDonnellLKE16,shirani2021learning}, and suggests that our IAA values mainly reflect the intrinsic ambiguity of the task rather than poorly specified instructions.

\section{Additional Details for Content Validity}
\label{app:content-validity}
We manually verify that the 120 \usecases for demonstrating the utility of \methodname in \S \ref{sec:content-validity} are constructed such that induced variations are lexically distinct. Assuming a very rigid relevance criterion of which deems that ``a \sample is relevant only if it explicitly states the induced variation'' (following which it is already checked for semantic relevance at the filtering stage) - even a lexical baseline is expected to give similar trends in the variation of the spread of relevant \samples across variations of the same usecase. 

\begin{figure}[htbp]
    \centering
    \includegraphics[width=0.9\linewidth]{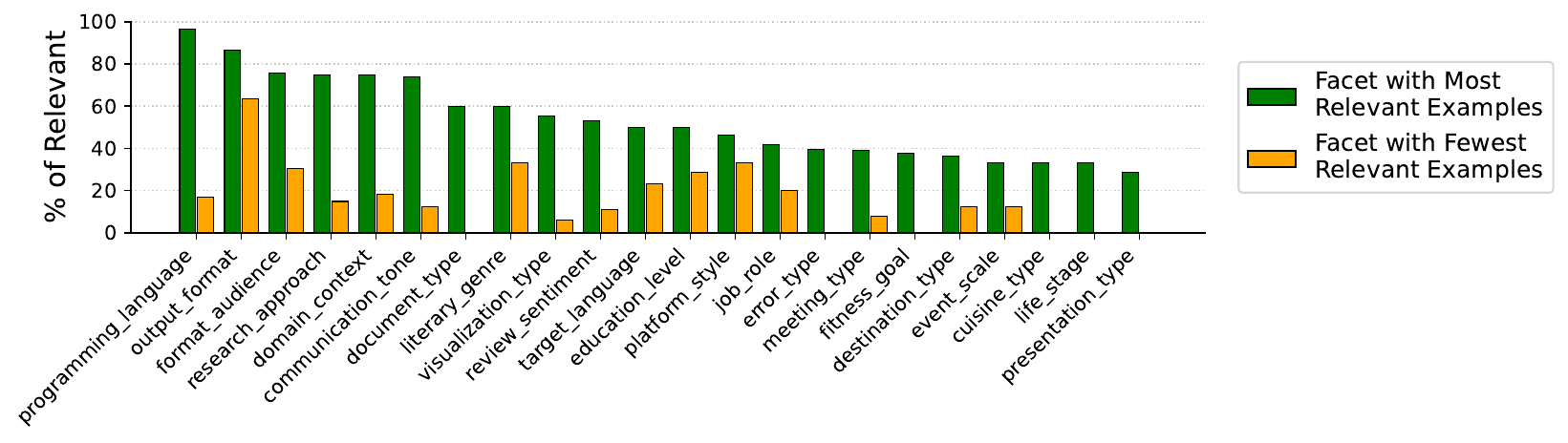}
    \caption{BM25 retrieval results for calibrating the variation in content validity as reported by \methodname}
    \label{fig:bm25-content-validity}
\end{figure}

Accordingly, we also construct a lexical BM25 baseline which only retrieves an \sample if it explicitly mentions the variation's keyword. 
Though stringent, this has been shown to be a realistic baseline of an annotator judging relevance for \usecases that they are either not interested in or are not experts in \citep{voorhees1998variations,Saracevic2007Relevance}. 
Figure~\ref{fig:bm25-content-validity} confirms that the qualitative trends from Figure~\ref{fig:content_validity} hold under this lexical baseline, suggesting that our findings are due to benchmark composition rather than an artifact of the semantic retriever.

We confirm that the lexical baseline also demonstrate the same trend in spread of relevance, reiterating that the variation in relevant count spread observed is an artifact of the properties of the \usecase and not due to lossy semantic retrieval. Ultimately, this confirms that some facets are underserved compared to other dominant facets.

\subsection{Example of Use Cases showing Spread in Relevant Count across Variations}

This table includes some examples of the task descriptions and consequent yields of \samples across variations along the same axes. 
\label{app:content-validity-examples}

\noindent\textbf{Table: Dispersion in Prec$_{sys}$@$k$ across facets of the same use-case.} 
Most use-cases show substantial spread: some variants (green) have many 
\textit{Relevant} examples, while others (red) are barely represented.
\label{tab:precision_dispersion}
\vspace{-2em}
\small
\begin{center}
\begin{tabular}{l r @{\hspace{2em}} l r @{\hspace{2em}} l r}
\toprule
\textbf{math word problems} & \textbf{Prec$_{sys}$@$k$} & \textbf{social media posts} & \textbf{Prec$_{sys}$@$k$} & \textbf{instruction following} & \textbf{Prec$_{sys}$@$k$} \\
\textit{Domain Context} &  & \textit{Platform Style} &  & \textit{Output Format} \\
\multicolumn{2}{l}{"solve finance calculation problem"} & \multicolumn{2}{l}{"write Instagram post"} & \multicolumn{2}{l}{"format response as Markdown"} \\
\midrule
finance calculation & \cellcolor{green!30}1.000 & Instagram & \cellcolor{green!30}0.862 & Markdown & \cellcolor{green!30}0.967 \\
travel distance & \cellcolor{green!30}0.933 & TikTok & \cellcolor{green!30}0.759 & JSON & \cellcolor{green!30}0.867 \\
construction material & \cellcolor{green!30}0.900 & Facebook & \cellcolor{green!30}0.714 & table & \cellcolor{yellow!30}0.633 \\
cooking measurement & \cellcolor{yellow!30}0.679 & Twitter & \cellcolor{green!30}0.700 & YAML & \cellcolor{orange!30}0.448 \\
inventory management & \cellcolor{yellow!30}0.500 & YouTube & \cellcolor{green!30}0.700 & bullet list & \cellcolor{orange!30}0.414 \\
&  & LinkedIn & \cellcolor{yellow!30}0.621 & &  \\
\bottomrule
\end{tabular}
\vspace{1em}
\begin{tabular}{l r @{\hspace{2em}} l r @{\hspace{2em}} l r}
\toprule
\textbf{creative writing} & \textbf{Prec$_{sys}$@$k$} & \textbf{summarization style} & \textbf{Prec$_{sys}$@$k$} & \textbf{email composition} & \textbf{Prec$_{sys}$@$k$} \\
\textit{Literary Genre} &  & \textit{Format Audience} &  & \textit{Communication Tone} \\
\multicolumn{2}{l}{"write fantasy story"} & \multicolumn{2}{l}{"summarize as legal memo"} & \multicolumn{2}{l}{"write follow-up email"} \\
\midrule
fantasy & \cellcolor{green!30}0.759 & legal memo & \cellcolor{green!30}1.000 & follow-up & \cellcolor{green!30}0.880 \\
science fiction & \cellcolor{yellow!30}0.679 & bullet points & \cellcolor{yellow!30}0.667 & formal business & \cellcolor{green!30}0.808 \\
mystery & \cellcolor{yellow!30}0.654 & executive summary & \cellcolor{yellow!30}0.533 & urgent request & \cellcolor{yellow!30}0.577 \\
romance & \cellcolor{yellow!30}0.625 & headline & \cellcolor{orange!30}0.483 & apologetic & \cellcolor{orange!30}0.476 \\
horror & \cellcolor{yellow!30}0.583 & tweet & \cellcolor{orange!30}0.483 & casual friendly & \cellcolor{orange!30}0.462 \\
historical fiction & \cellcolor{yellow!30}0.517 & academic abstract & \cellcolor{orange!30}0.440 & thank you & \cellcolor{orange!30}0.391 \\
\bottomrule
\end{tabular}
\vspace{1em}
\begin{tabular}{l r @{\hspace{2em}} l r @{\hspace{2em}} l r}
\toprule
\textbf{debugging code} & \textbf{Prec$_{sys}$@$k$} & \textbf{data analysis} & \textbf{Prec$_{sys}$@$k$} & \textbf{research methodology} & \textbf{Prec$_{sys}$@$k$} \\
\textit{Error Type} &  & \textit{Visualization Type} &  & \textit{Research Approach} \\
\multicolumn{2}{l}{"debug syntax error"} & \multicolumn{2}{l}{"create line graph analysis"} & \multicolumn{2}{l}{"design experimental study"} \\
\midrule
syntax error & \cellcolor{green!30}0.867 & line graph & \cellcolor{yellow!30}0.593 & experimental study & \cellcolor{green!30}0.759 \\
runtime error & \cellcolor{green!30}0.714 & bar chart & \cellcolor{yellow!30}0.556 & quantitative research study & \cellcolor{yellow!30}0.655 \\
logic error & \cellcolor{yellow!30}0.690 & scatter plot & \cellcolor{yellow!30}0.500 & observational study & \cellcolor{yellow!30}0.552 \\
security vulnerability & \cellcolor{orange!30}0.393 & pie chart & \cellcolor{yellow!30}0.500 & case study research & \cellcolor{orange!30}0.407 \\
performance issue & \cellcolor{orange!30}0.357 & histogram & \cellcolor{orange!30}0.481 & qualitative research study & \cellcolor{orange!30}0.357 \\
memory leak & \cellcolor{red!30}0.269 & heatmap & \cellcolor{orange!30}0.333 & mixed methods study & \cellcolor{red!30}0.208 \\
\bottomrule
\end{tabular}
\vspace{1em}
\begin{tabular}{l r @{\hspace{2em}} l r @{\hspace{2em}} l r}
\toprule
\textbf{code generation} & \textbf{Prec$_{sys}$@$k$} & \textbf{language translation} & \textbf{Prec$_{sys}$@$k$} & \textbf{event planning} & \textbf{Prec$_{sys}$@$k$} \\
\textit{Programming Language} &  & \textit{Target Language} &  & \textit{Event Scale} \\
\multicolumn{2}{l}{"write function in Python"} & \multicolumn{2}{l}{"translate text to Chinese"} & \multicolumn{2}{l}{"plan small dinner party"} \\
\midrule
Python & \cellcolor{green!30}0.867 & Chinese & \cellcolor{yellow!30}0.600 & small dinner party & \cellcolor{yellow!30}0.600 \\
C++ & \cellcolor{green!30}0.741 & German & \cellcolor{yellow!30}0.586 & birthday celebration & \cellcolor{yellow!30}0.524 \\
Rust & \cellcolor{orange!30}0.478 & French & \cellcolor{orange!30}0.467 & large conference & \cellcolor{orange!30}0.414 \\
JavaScript & \cellcolor{orange!30}0.346 & Japanese & \cellcolor{orange!30}0.448 & fundraising gala & \cellcolor{orange!30}0.412 \\
Java & \cellcolor{red!30}0.286 & Spanish & \cellcolor{orange!30}0.379 & corporate retreat & \cellcolor{orange!30}0.333 \\
Go & \cellcolor{red!30}0.100 & Arabic & \cellcolor{red!30}0.276 & &  \\
\bottomrule
\end{tabular}
\vspace{1em}
\begin{tabular}{l r @{\hspace{2em}} l r @{\hspace{2em}} l r}
\toprule
\textbf{trip planning} & \textbf{Prec$_{sys}$@$k$} & \textbf{financial advice} & \textbf{Prec$_{sys}$@$k$} & \textbf{presentation slides} & \textbf{Prec$_{sys}$@$k$} \\
\textit{Destination Type} &  & \textit{Life Stage} &  & \textit{Presentation Type} \\
\multicolumn{2}{l}{"plan trip to multiple cities"} & \multicolumn{2}{l}{"provide financial advice for you..."} & \multicolumn{2}{l}{"create conference presentation s..."} \\
\midrule
trip to multiple cities & \cellcolor{green!30}0.733 & young professional & \cellcolor{yellow!30}0.593 & conference & \cellcolor{yellow!30}0.619 \\
trip under budget cap & \cellcolor{yellow!30}0.640 & mid-career adult & \cellcolor{yellow!30}0.538 & business & \cellcolor{orange!30}0.483 \\
trip to Asia & \cellcolor{orange!30}0.467 & pre-retiree & \cellcolor{yellow!30}0.500 & academic & \cellcolor{orange!30}0.464 \\
trip to Europe & \cellcolor{orange!30}0.400 & retiree & \cellcolor{yellow!30}0.500 & sales & \cellcolor{orange!30}0.458 \\
trip to rural village & \cellcolor{red!30}0.267 & college student & \cellcolor{orange!30}0.333 & investor & \cellcolor{orange!30}0.455 \\
weekend trip only & \cellcolor{red!30}0.227 & new parent & \cellcolor{red!30}0.263 & training & \cellcolor{red!30}0.222 \\
\bottomrule
\end{tabular}
\vspace{1em}
\begin{tabular}{l r @{\hspace{2em}} l r @{\hspace{2em}} l r}
\toprule
\textbf{product reviews} & \textbf{Prec$_{sys}$@$k$} & \textbf{meeting notes} & \textbf{Prec$_{sys}$@$k$} & \textbf{lesson planning} & \textbf{Prec$_{sys}$@$k$} \\
\textit{Review Sentiment} &  & \textit{Meeting Type} &  & \textit{Education Level} \\
\multicolumn{2}{l}{"write mixed product review"} & \multicolumn{2}{l}{"take project meeting notes"} & \multicolumn{2}{l}{"create preschool lesson plan"} \\
\midrule
mixed & \cellcolor{yellow!30}0.552 & project meeting & \cellcolor{yellow!30}0.565 & preschool & \cellcolor{yellow!30}0.579 \\
negative & \cellcolor{yellow!30}0.500 & board meeting & \cellcolor{yellow!30}0.526 & elementary school & \cellcolor{orange!30}0.458 \\
detailed & \cellcolor{orange!30}0.370 & team standup & \cellcolor{orange!30}0.478 & college & \cellcolor{orange!30}0.407 \\
brief & \cellcolor{red!30}0.286 & client meeting & \cellcolor{orange!30}0.409 & high school & \cellcolor{orange!30}0.333 \\
&  & interview meeting & \cellcolor{orange!30}0.333 & middle school & \cellcolor{orange!30}0.320 \\
&  & brainstorming meeting & \cellcolor{red!30}0.240 & adult education & \cellcolor{orange!30}0.304 \\
\bottomrule
\end{tabular}
\vspace{1em}
\begin{tabular}{l r @{\hspace{2em}} l r @{\hspace{2em}} l r}
\toprule
\textbf{interview questions} & \textbf{Prec$_{sys}$@$k$} & \textbf{legal documents} & \textbf{Prec$_{sys}$@$k$} & \textbf{workout design} & \textbf{Prec$_{sys}$@$k$} \\
\textit{Job Role} &  & \textit{Document Type} &  & \textit{Fitness Goal} \\
\multicolumn{2}{l}{"create sales representative inte..."} & \multicolumn{2}{l}{"draft partnership agreement"} & \multicolumn{2}{l}{"design workout for rehabilitation"} \\
\midrule
sales representative & \cellcolor{yellow!30}0.500 & partnership agreement & \cellcolor{orange!30}0.385 & rehabilitation & \cellcolor{yellow!30}0.500 \\
data scientist & \cellcolor{orange!30}0.455 & licensing agreement & \cellcolor{orange!30}0.370 & weight loss & \cellcolor{red!30}0.273 \\
marketing manager & \cellcolor{orange!30}0.316 & employment contract & \cellcolor{orange!30}0.333 & strength & \cellcolor{red!30}0.261 \\
customer service & \cellcolor{orange!30}0.300 & non-disclosure agreement & \cellcolor{red!30}0.250 & muscle gain & \cellcolor{red!30}0.208 \\
software engineer & \cellcolor{red!30}0.292 & privacy policy & \cellcolor{red!30}0.231 & endurance & \cellcolor{red!30}0.185 \\
product manager & \cellcolor{red!30}0.200 & terms of service & \cellcolor{red!30}0.182 & flexibility & \cellcolor{red!30}0.111 \\
\bottomrule
\end{tabular}
\end{center}
% \begin{tabular}{l r}
% \toprule
% \textbf{recipe creation} & \textbf{Prec$_{sys}$@$k$} \\
% \textit{Cuisine Type} \\
% \multicolumn{2}{l}{"create Indian recipe"} \\
% \midrule
% Indian & \cellcolor{orange!30}0.333 \\
% Thai & \cellcolor{orange!30}0.333 \\
% Chinese & \cellcolor{red!30}0.278 \\
% Mexican & \cellcolor{red!30}0.250 \\
% Italian & \cellcolor{red!30}0.182 \\
% French & \cellcolor{red!30}0.111 \\
% \bottomrule
% \end{tabular}

\section{Additional Details for Convergent Validity}
\label{app:convergent-validity}

We use the following eight models for our evaluation: \texttt{Llama-3.1-8B-Instruct}, \texttt{Llama-3.1-Tulu-3-8B}, \texttt{Mistral-7B-Instruct-v0.2}, \texttt{Qwen1.5-14B-Chat}, \texttt{Qwen2.5-7B-Instruct}, \texttt{c4ai-command-r7b-12-2024}, \texttt{gemma-2-9b-it}, and \texttt{gemma-3-4b-it}.

\paragraph{Accuracy Aggregation for Model Performance Estimates}
\label{app:benchmark-metrics}
Model performance is evaluated using the specific metric designated by each benchmark, as summarized in Table \ref{tab:benchmark-metrics}. The majority of benchmarks supported by the Eval Harness \cite{eval-harness} provide per-sample accuracy based on strict string matching or binary accuracy. We utilize these values directly, noting that they are bounded between 0 and 1 (representing a binary exact match or correct prediction).For benchmarks employing Win-Rate as a metric, we generate model inferences and compute the win-rate against a fixed set of standard responses generated by the \texttt{Ministral-8B-Instruct-2410} model. In this comparison, a win against the reference model is encoded as 1, while a tie or loss is encoded as 0 (serving as an accuracy equivalent).For the remaining benchmarks requiring custom evaluation protocols, we employ the specific metrics and libraries provided by the original authors. For instance, InfoBench \cite{qin2024infobench} is evaluated using the provided code for its designated DRFR metric. These per-sample scores are subsequently aggregated to derive the final model performance estimates.

\begin{table}[h]
\centering
\caption{Summary of evaluation metrics used across benchmarks.}
\begin{tabular}{l p{0.25\linewidth} p{0.55\linewidth}}
\toprule
\textbf{Category} & \textbf{Key Used} & \textbf{Processors / Benchmarks} \\
\midrule
\textbf{Custom Metric} & \textit{Varied} & complexbench (point\_judges), infobench (DRFR), ifeval (inst\_level\_strict\_acc), llmbar (gold label) \\
\midrule
\textbf{Accuracy} & acc & nutribench, truthfulQA, arc, mmlu, winogrande, hellaswag, headqa, logiqa, mathqa, openbookqa, sciq, pubmed, wmdp, moral\_stories, big-bench tasks \\
 & exact\_match & triviaqa \\
 & pass@1 & humaneval \\
\midrule
\textbf{Exact Match} & exact\_match & bbh, gsm8k, webqs \\
\midrule
\textbf{F1 Score} & f1/f1\_abstractive & qasper \\
\midrule
\textbf{Win Rate} & win & mtbench, koala, vicuna, lfqa, alpacaeval \\
\bottomrule
\end{tabular}
\label{tab:benchmark-metrics}
\end{table}

\label{app:convergent-validity-additional-details}

To control for any variance in retrieval results induced due to difference in prompting language - we write 5 task descriptions per \usecase and retrieve \samples for each. Then we pick the prompt that has the high automatic precision and have those annotated by human annotators. To account for the change the Kendall's $\tau$ due to the size of the retrieved set and measure the internal heterogeneity of the gold set, we compare the Kendall's $\tau$ across 3 configurations which is described in detail below. 

\subsection{Correcting Kendall's Tau}

For each task, we first select a single prompt per (model, task): the prompt that maximizes the model’s mean accuracy over that prompt’s retrieved subset.
Let $a^{\mathrm{ret}}_m$ denote model $m$’s mean accuracy on its retrieved items (with $n_m$ retrieved items), and let $\bar{a}^{\mathrm{gold}}_m$ denote its mean accuracy on the full gold test set.
We report three Kendall’s $\tau$ correlations per task (averaged over $S$ Monte Carlo trials):

\begin{enumerate}[leftmargin=*, label=(\roman*)]
    \item \textbf{Same-cardinality retrieved--gold: $\tau_{\mathrm{ret}}$.}
    We compare the retrieved ranking $\{a^{\mathrm{ret}}_m\}_m$ to a gold ranking estimated at \emph{matched evaluation budget}.
    In each trial, for every model $m$ we subsample $n_m$ examples from its gold pool, compute the subsampled gold accuracy $a^{\mathrm{gold},(s)}_m$, and compute
    $\tau\!\left(\{a^{\mathrm{ret}}_m\}_m,\{a^{\mathrm{gold},(s)}_m\}_m\right)$. $\tau_{\mathrm{ret}}$ is meant to ascertain if the retrieved set induces the same model ordering as the gold set, if both were evaluated with the same number of examples. 

    \item \textbf{Internal gold stability (full vs.\ subset): $\tau_{\mathrm{gold}}$.}
    To quantify how much rankings can fluctuate due to \emph{finite-sample effects alone}, we measure stability of gold rankings when reducing the gold set to $n$ examples.
    Let $n=\lfloor \frac{1}{M}\sum_m n_m \rfloor$ be the mean retrieved cardinality for the task.
    In each trial we subsample $n$ gold examples per model, compute $a^{\mathrm{gold},(s)}_m$, and compute
    $\tau\!\left(\{\bar{a}^{\mathrm{gold}}_m\}_m,\{a^{\mathrm{gold},(s)}_m\}_m\right)$.
    Thus, $\tau_{\mathrm{gold}}$ provides a task-specific ceiling on rank agreement attributable to reduced evaluation budget.

    \item \textbf{Gold--gold repeatability (subset vs.\ subset): $\tau_{\mathrm{sanity}}$.}
    As an additional control, we measure ranking repeatability when \emph{both} rankings are estimated from gold subsets of the same size.
    In each trial, we draw two independent subsamples of size $n$ per model, compute the corresponding accuracies, and compute
    $\tau\!\left(\{a^{\mathrm{gold},(s_1)}_m\}_m,\{a^{\mathrm{gold},(s_2)}_m\}_m\right)$.
    This isolates the intrinsic uncertainty in rankings when the evaluation budget is $n$ (without referencing the full gold ranking).
\end{enumerate}

\begin{figure}
    \centering
    \small
    \includegraphics[width=0.5\linewidth]{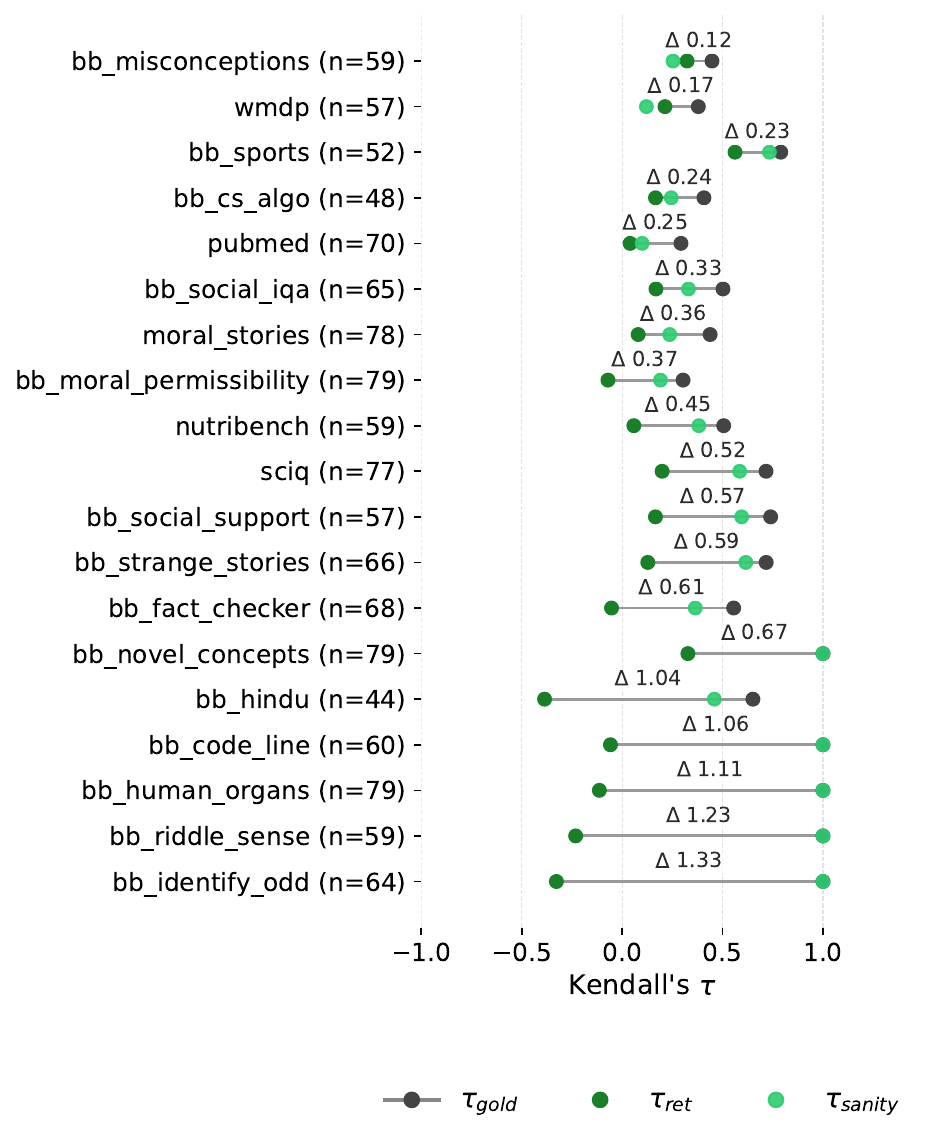}
    \caption{$\tau_{\mathrm{sanity}}$ and $\tau_{\mathrm{gold}}$ provide gold-only stability controls; $\tau_{\mathrm{ret}}$ measures additional divergence introduced by retrieval.}
    \label{fig:convergent-tau-comparison-detailed}
\end{figure}

Across tasks (Fig.\ref{fig:convergent-tau-comparison-detailed}), $\tau_{\mathrm{gold}}$ and $\tau_{\mathrm{sanity}}$ quantify how reliable model ordering is even within the original benchmark at the relevant sample size. Across most tasks, the $\tau_{\text{sanity}}$ are relatively close to the maximum achievable $\tau_{gold}$, indicating that model rankings are \emph{stable} at the retrieved-set cardinality \textit{internally}; This helps us confirm that the low $\tau_{ret}$ we observe is not due to inherent heterogeneity in the gold set. 
However, $\tau_{\text{ret}}$ is often much lower (sometimes near-zero or negative) indicative of alternate explanations: we posit, divergent operationalizations and rubric mismatches which ultimately engender claims with low convergent validity for those \usecases.

\section{Limitations}
Our human evaluation is constrained in scale (number and diversity of \usecases and annotators) to keep annotation cost manageable: while we collect 4,200 relevance judgments, precision and coverage estimates may still not fully generalize beyond the evaluated \usecases.

The content-validity analysis in \S\ref{sec:content-validity} does not elicit or verify facets from annotators, avoiding a second layer of costly human validation; Additionally, these facets are generated by a language model, which introduces risk of biased recall of facets. We expect that using the LLM as Judge can provide a reasonable preliminary signal of the any representational disparity for a \usecase which can be significantly improved using human supervision. Consequently, analysis in this section is intended to be useful primarily for highlighting the presence of representation disparities that practitioners can inspect rather than as actionable claims about which facets must be served more equitably since that is heavily conditioned on a human-in-the-loop elicitation and relevance review assessment of the retrieved \samples.

Finally, \methodname can only surface misalignment within the benchmark items and model outputs it indexes; expanding coverage (more benchmark suites and models, or commercial-model inferences required for some metrics) is resource-intensive, so any diagnosis is bounded by the current index and the feasible set of evaluated models.

% You can have as much text here as you want. The main body must be at most $8$
% pages long. For the final version, one more page can be added. If you want, you
% can use an appendix like this one.

% The $\mathtt{\backslash onecolumn}$ command above can be kept in place if you
% prefer a one-column appendix, or can be removed if you prefer a two-column
% appendix.  Apart from this possible change, the style (font size, spacing,
% margins, page numbering, etc.) should be kept the same as the main body.
%%%%%%%%%%%%%%%%%%%%%%%%%%%%%%%%%%%%%%%%%%%%%%%%%%%%%%%%%%%%%%%%%%%%%%%%%%%%%%%
%%%%%%%%%%%%%%%%%%%%%%%%%%%%%%%%%%%%%%%%%%%%%%%%%%%%%%%%%%%%%%%%%%%%%%%%%%%%%%%

\end{document}